\documentclass[sigplan,screen,nonacm]{acmart}

\setcopyright{none}
\copyrightyear{2027}
\acmYear{2027}
\acmConference[ASPLOS '27]{Proceedings of the 32nd ACM International Conference on Architectural Support for Programming Languages and Operating Systems, Volume 1}{April 11--15, 2027}{Heraklion, Greece}
\acmBooktitle{Proceedings of the 32nd ACM International Conference on Architectural Support for Programming Languages and Operating Systems, Volume 1 (ASPLOS '27), April 11--15, 2027, Heraklion, Greece}
\acmDOI{10.1145/3845814.3850026}
\acmISBN{979-8-4007-3012-2/2027/04}

\makeatletter
\AtBeginDocument{%
  \ifx\acmConference@shortname\@empty
    \let\footnotetextcopyrightpermission\@gobble
  \fi
}
\makeatother

\AtBeginDocument{%
  }

\usepackage{algorithm}
\usepackage{algorithmic}
\usepackage{subcaption}
\usepackage{multirow}
\usepackage{makecell}
\usepackage{array}
\usepackage{fancyvrb}
\usepackage{balance}

\title[Specter]{Specter: Fast MoE Inference via Quantization-Enabled Multi-Step Speculative Offloading}

\author{Wenfeng Wang}
\email{wangwenfeng@sjtu.edu.cn}
\affiliation{%
  \institution{Shanghai Jiao Tong University}
  \city{Shanghai}
  \country{China}
}

\author{Xiaofeng Hou}
\authornote{Chao Li and Xiaofeng Hou are the corresponding authors.}
\email{hou-xf@cs.sjtu.edu.cn}
\affiliation{%
  \institution{Shanghai Jiao Tong University}
  \city{Shanghai}
  \country{China}
}

\author{Xinfeng Xia}
\email{xf6767@sjtu.edu.cn}
\affiliation{%
  \institution{Shanghai Jiao Tong University}
  \city{Shanghai}
  \country{China}
}

\author{Xinkai Wang}
\email{unbreakablewxk@sjtu.edu.cn}
\affiliation{%
  \institution{Shanghai Jiao Tong University}
  \city{Shanghai}
  \country{China}
}

\author{Mingxuan Zhang}
\email{mingxuanzhang@sjtu.edu.cn}
\affiliation{%
  \institution{Shanghai Jiao Tong University}
  \city{Shanghai}
  \country{China}
}

\author{Peng Tang}
\email{tttppp@sjtu.edu.cn}
\affiliation{%
  \institution{Shanghai Jiao Tong University}
  \city{Shanghai}
  \country{China}
}

\author{Jiacheng Liu}
\email{liujiacheng@sjtu.edu.cn}
\affiliation{%
  \institution{Shanghai Jiao Tong University}
  \city{Shanghai}
  \country{China}
}

\author{Chao Li}
\authornotemark[1]
\email{lichao@cs.sjtu.edu.cn}
\affiliation{%
  \institution{Shanghai Jiao Tong University}
  \city{Shanghai}
  \country{China}
}

\renewcommand{\shortauthors}{Wenfeng Wang et al.}

\begin{document}

\begin{abstract}

Mixture-of-experts (MoE) models often exceed a single accelerator's memory capacity. Offloading experts to host memory creates a severe PCIe I/O bottleneck because data-dependent routing keeps transfers on the critical path.

We present \textsc{Specter}, an inference system that uses quantization to combine speculative decoding with expert offloading. Its lightweight quantized draft model runs efficiently on device while preserving high-fidelity expert activation patterns. This provides accurate multi-step lookahead over future expert usage, allowing the runtime to organize expert movement and execution ahead of verification and improve both communication and execution efficiency.

Under single-request inference, \textsc{Specter} achieves an average $1.50\times$ (up to $1.88\times$) speedup in time per output token (TPOT) over the state-of-the-art system. \textsc{Specter} makes memory-constrained, single-GPU MoE inference more practical on commodity hardware. Our code is available at \url{https://github.com/wwwwwf/specter_asplos}.

\end{abstract}


\maketitle

\makeatother

\section{Introduction}
\label{sec:introduction}

\begin{figure}[!htbp]
    \centering
    \includegraphics[width=\columnwidth]{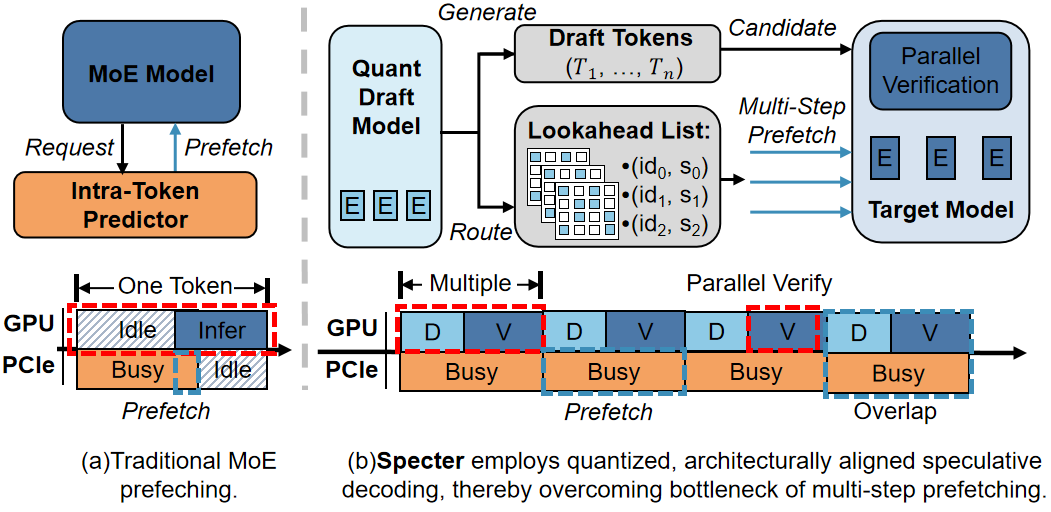} 
    \caption{Conventional offloading versus \textsc{Specter}. Aligned quantized speculative decoding lays the foundation for accurate prefetching and efficient system design.}
    \Description{Two diagrams and timelines contrast intra-token prefetching with multi-step speculative offloading. The conventional path alternates PCIe transfers and GPU inference, leaving idle intervals. The Specter path uses a quantized draft model to produce candidate tokens and an expert lookahead list, allowing expert prefetching to overlap with drafting and target verification.}
    \label{fig:first}
\end{figure}

The mixture-of-experts (MoE) architecture~\cite{shazeer2017moe} has become the de facto standard for scaling large language models (LLMs). By sparsely routing every token to different parameter subsets, MoE models such as Mixtral-8x7B~\cite{Jiang2024Mixtral}, Phi-MoE~\cite{abdin2024phi3technicalreporthighly}, Qwen-MoE~\cite{yang2025qwen3}, and DeepSeek~\cite{guo2025deepseek} can achieve superior quality without any increase in computational cost. However, its enormous memory footprint presents a fundamental deployment challenge, often exceeding the capacity of a single accelerator. Offloading inactive experts to host memory~\cite{eliseev2023fast, sheng2023flexgen, cao2025moe} is a pragmatic solution, which shifts the system bottleneck from computation to data movement. Understanding end-to-end execution bottlenecks on GPUs is crucial for optimizing modern neural networks~\cite{9924614, xu2021dataclue}. In this regard, GPU execution is frequently stalled by PCIe transfers, severely inflating end-to-end inference latency.

A natural response to mitigate this PCIe latency is expert prefetching~\cite{zhong2024adapmoe,song2024promoe,tang2026moe}. However, the effectiveness of prefetching in MoE inference is fundamentally limited by the autoregressive nature of generation. Because routing decisions are highly input-dependent and strictly sequential, existing prefetching methods are largely restricted to intra-token prediction, i.e., leveraging hidden states from early layers to predict later-layer routes for the current token. This approach only yields a microsecond-level computation window, which is fundamentally insufficient to overlap with the millisecond-level PCIe transfer latency of gigabyte-sized experts. Consequently, the GPU remains starved, and significant stalls remain unresolved. To truly hide I/O latency, the system requires multi-step, inter-token lookahead.

This observation motivates our key insight: \textbf{speculative decoding (SD) naturally provides the multi-step lookahead required for aggressive I/O scheduling} (Figure~\ref{fig:first}). Traditionally, SD employs a lightweight, independent draft model to rapidly generate future tokens. However, directly applying this paradigm to MoE offloading reveals a critical mismatch: a standard tiny dense draft model, while extremely memory-efficient, lacks the routing mechanism entirely, making it incapable of predicting \textit{which} experts the target MoE will activate. Conversely, training a smaller independent MoE draft model incurs significant cost and often suffers from routing misalignment with the target model. To resolve this dilemma, we discover that a low-bit quantized version (e.g., 4-bit integer (INT4)) of the target MoE model itself serves as an ideal, zero-training-cost predictor. By sharing the identical routing architecture and parameter lineage, this quantized draft guarantees high-fidelity expert activation predictions. As illustrated in Figure~\ref{fig:first}, \textsc{Specter} leverages this quantized draft to create a sufficiently large multi-step time window, orchestrating heavy PCIe transfers well in advance and effectively hiding the I/O latency behind useful computation. Meanwhile, the rigorous draft-then-verify mechanism mathematically guarantees zero degradation in generation quality and output distribution.

Turning this intuition into a high-performance system, however, exposes severe system-level challenges. Applying SD to MoE models triggers an "expert explosion". In dense models, the verification cost of SD scales gracefully, growing only marginally as the speculation depth $K$ increases. In contrast, predicting $K$ future tokens in an MoE model simultaneously activates a highly diverse set of experts. This sudden burst of demand instantly saturates PCIe bandwidth and overwhelms limited GPU memory. Furthermore, the system must carefully orchestrate asynchronous pipelines—draft generation, expert prefetching, and target verification—without introducing synchronization bubbles. A naive integration of SD and MoE offloading simply replaces computation stalls with severe I/O thrashing.
While recent works have begun exploring SD for MoE inference, they largely introduce explicit quality-latency tradeoffs~\cite{mcdanel2026moe,10.1145/3774904.3792218}, or require structurally compatible draft--target model pairs~\cite{chen2025sp}, leaving scheduling and memory wall challenges unresolved.

To overcome these challenges, we design and build \textsc{Specter}, a co-designed MoE 
inference acceleration system. {We target memory-constrained MoE inference on a single GPU, with model weights partially offloaded to host memory.} \textsc{Specter} uniquely leverages SD to deliver a dual-acceleration paradigm: it utilizes the draft model's multi-step lookahead to aggressively prefetch experts and hide millisecond-level PCIe I/O latency, while simultaneously exploiting SD's parallel verification to accelerate the compute-bound generation phase. 
In summary, \textsc{Specter} consists of the following three core components:
\begin{itemize}
\item \textbf{Speculative Inference Controller (SIC):} We propose a novel MoE offloading framework that integrates a quantized draft model to provide high-fidelity, multi-step lookahead. The system selects the speculation depth through offline profiling to balance I/O overhead and compute acceleration.
\item \textbf{Predictive I/O Orchestrator (PIO):} To resolve the "expert explosion" bottleneck, we design a \emph{Prefetch Planner} guided by an Expert Lookahead Buffer (ELB). It translates draft predictions into precise, asynchronous PCIe prefetching and cache management, effectively hiding I/O latency behind computation.
\item \textbf{Streamlined Execution Engine (SEE):} We develop a highly optimized execution backend that accelerates both phases of SD. It features fused quantized MoE kernels for rapid draft generation and expert-centric computation reordering to streamline target-side verification, eliminating pipeline synchronization bubbles.
\end{itemize}

We implement \textsc{Specter} and evaluate it across three representative MoE architectures under strict hardware constraints. 
Compared with SpecMoEOff~\cite{wang2025accelerating}, our primary MoE speculative-offloading
baseline, \textsc{Specter} achieves an average 1.50$\times$ TPOT
speedup and up to 1.88$\times$ under single-request inference.

\section{Background and Motivation}
\label{sec:background}

\subsection{MoE Offloading Demands Multi-step Lookahead}

Recently, MoE models have been shown to scale model capacity without increasing computational cost, thus providing higher accuracy~\cite{shazeer2017moe,Jiang2024Mixtral,aminabadi2022deepspeed}. An MoE layer replaces the standard dense feed-forward network (FFN) with a set of $N$ "expert" networks and a trainable router. For each input token, the router dynamically selects a small subset of experts (e.g., top-2) to process the token's hidden state. Despite its advantages, the MoE inference system has to offload~\cite{eliseev2023fast, sheng2023flexgen, cao2025moe} the bloat of expert parameters to host memory and load them as needed for inference. This leads to massive synchronous stalls at every MoE layer.

Recent works attempt to mitigate these GPU stalls through expert prefetching. Nevertheless, because MoE routing is highly dependent on the current token's hidden states, existing methods are largely restricted to intra-token prediction. They leverage hidden states from earlier layers to predict later-layer routes for the current token. These approaches only yield a microsecond-level computation window, which is fundamentally insufficient to overlap with the millisecond-level PCIe transfer latency of gigabyte-scale expert working sets. Extending to multi-step lookahead without a verification mechanism is impractical; forcing the target model to execute based on unverified future predictions inevitably accumulates routing errors and degrades the final generation quality. Consequently, significant GPU stalls remain unresolved in current offloading systems.

\subsection{Speculative Decoding Provides Multi-Step Lookahead}
\label{ssec:spec_dec_background}

Recently, speculative decoding~\cite{leviathan2023fast, chen2023accelerating} offers a principled solution to this dilemma by decoupling prediction from final execution. In SD, a lightweight draft model rapidly generates a sequence of $K$ future tokens, which the target model then verifies in parallel. This accept/reject mechanism mathematically guarantees lossless generation quality, even if the draft predictions are imperfect. For MoE offloading, the SD framework naturally provides a multi-step time window to orchestrate PCIe prefetching. The drafting stage, running entirely in GPU VRAM, races ahead to generate future tokens. We can extract future expert routing information during this stage to orchestrate PCIe prefetching. Even if the draft model occasionally mispredicts, the verification stage guarantees that the final output distribution remains completely lossless.

In traditional speculative decoding, a smaller, independent model from the same family is typically employed as the draft. However, applying this paradigm to MoE offloading presents fundamental challenges. First, \textit{lack of routing correspondence}: a smaller dense model lacks an expert routing mechanism entirely, and even a smaller MoE model possesses a different routing space (e.g., different hidden dimensions or expert counts), making it impossible to directly map its expert activations to the target model. Second, \textit{poor universality and availability}: not all open-source MoE families provide a perfectly matched, lightweight counterpart, and training a dedicated draft model incurs prohibitive costs. Third, \textit{memory redundancy}: an independent draft model requires maintaining separate weights and a redundant key–value (KV) cache, which severely exacerbates the memory wall and leaves less VRAM for caching experts.

To overcome these hurdles and maximize the effectiveness of this speculative prefetching, the draft model must foresee the target's expert routing with high accuracy while minimizing additional memory overhead. Fortunately, we find that a quantized version of the target MoE itself (e.g., INT4) serves as an ideal draft predictor. Because it shares the identical routing architecture and parameter lineage, it provides high-fidelity routing predictions without any training overhead. To demonstrate this, we measured the expert selection alignment between a full-precision 16-bit floating-point (FP16) Qwen1.5-MoE (the target) and its INT4 quantized counterpart (the draft). For each token, we compared the top-4 experts selected by the draft against those selected by the target.

\begin{figure}[!htbp]
    \centering
    \includegraphics[width=0.45\textwidth]
    {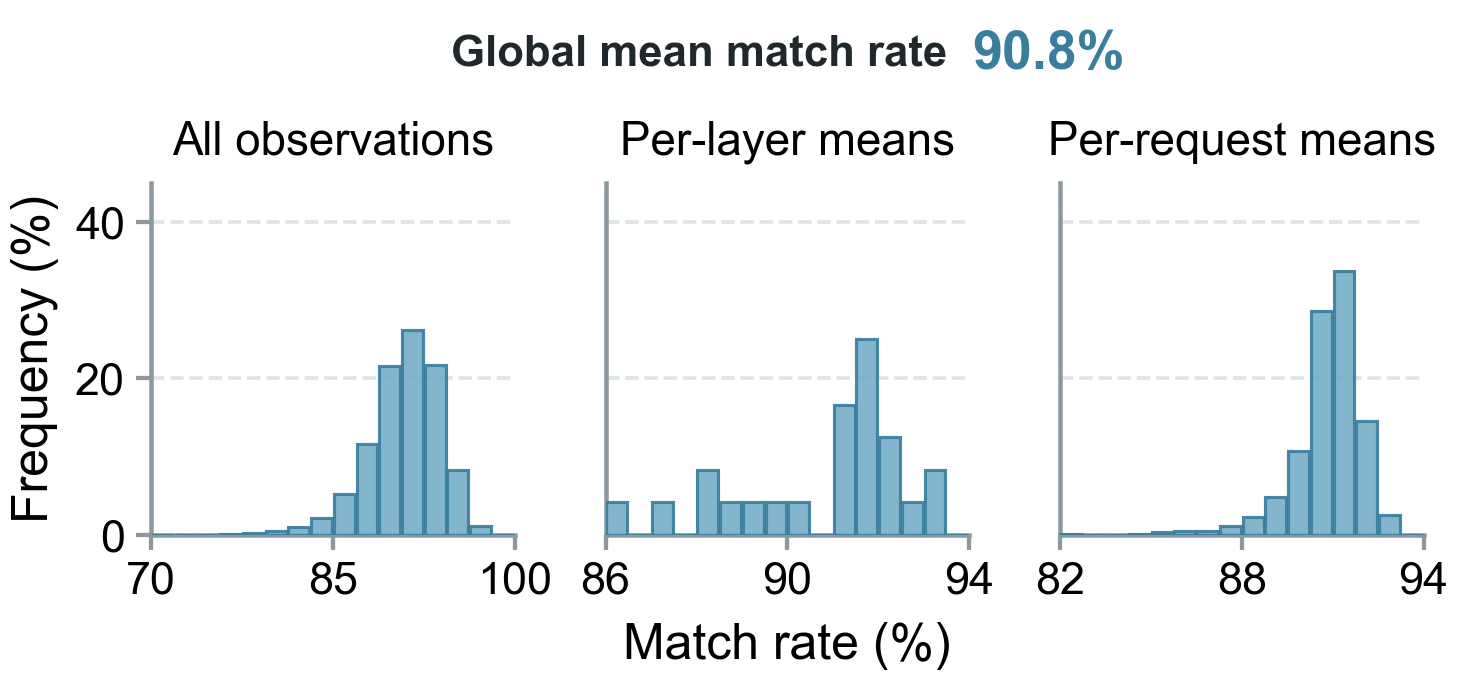}
    \caption{{Expert-ID selection fidelity between an INT4 Qwen1.5-MoE
    draft model and its FP16 target model.}}
    \Description{Three histograms show expert-ID match-rate distributions for all observations, per-layer means, and per-request means. The horizontal axes show match rate and the vertical axis shows frequency, both in percent. All three distributions concentrate around 90 to 92 percent, with a reported global mean of 90.8 percent.}
    \label{fig:total_match}
\end{figure}

\begin{table}[!htbp]
\centering
\caption{Expert activation similarity (\%) across MoE models
(using native top-$k$ configuration and GPTQ method), diverse
quant methods~\cite{frantar2022gptq, lin2024awq}, and GGUF-format precisions (based on DeepSeekV2-Lite).}
\label{tab:fidelity}

\begingroup
\setlength{\tabcolsep}{4pt}
\renewcommand{\arraystretch}{1.2}
\resizebox{\columnwidth}{!}{%
\begin{tabular}{|c|ccc|ccc|ccc|}
\hline
\multirow{2}{*}{} &
\multicolumn{3}{c|}{\textbf{Model@(GPTQ\&Q4)}} &
\multicolumn{3}{c|}{\textbf{Method@(DS\&Q4)}} &
\multicolumn{3}{c|}{\textbf{Precision@(GGUF\&DS)}} \\
\cline{2-10}
& \textbf{Phi} & \textbf{DS} & \textbf{Qwen}
& \textbf{GGUF} & \textbf{GPTQ} & \textbf{AWQ}
& \textbf{Q2} & \textbf{Q4} & \textbf{Q8} \\
\hline
\textbf{Max}
& 99.79 & 99.45 & 93.48
& 99.99 & 99.45 & 99.45
& 99.98 & 99.99 & 99.99 \\
\hline
\textbf{Min}
& 95.15 & 94.49 & 86.30
& 95.09 & 94.49 & 93.17
& 90.10 & 95.09 & 98.66 \\
\hline
\textbf{Mean}
& \textbf{\textit{96.50}} & \textbf{\textit{96.33}} & \textbf{\textit{90.82}}
& \textbf{\textit{96.53}} & \textbf{\textit{96.33}} & \textbf{\textit{95.72}}
& \textbf{\textit{92.96}} & \textbf{\textit{96.53}} & \textbf{\textit{99.38}} \\
\hline
\end{tabular}%
}
\endgroup
\end{table}

{{
Figure~\ref{fig:total_match} analyzes the top-4 expert-ID match rates between the INT4 draft and FP16 target models for Qwen1.5-MoE across 1,000 requests and 24 MoE layers (24,000 request–layer pairs). To dissect the stability of this lookahead, we evaluate the match rate across three dimensions: the raw instance-level distribution, the layer-wise aggregation (evaluating architectural depth sensitivity), and the request-wise aggregation (evaluating workload variability). All three distributions consistently peak above 90\% with a global mean of 90.8\%, confirming that INT4 quantization preserves high-fidelity expert routing across evaluated layers and inputs.
Table~\ref{tab:fidelity} further confirms this trend across models, quantization methods, and precision settings.}
}

This high-fidelity, zero-training-cost predictability is transformative: it allows us to convert the highly non-deterministic, data-dependent expert routing problem into a deterministic I/O prefetching and scheduling problem, enabling the orchestration of heavy PCIe transfers well in advance.

\subsection{Speculative Offloading Needs System Efforts}

\begin{figure}[!htbp] 
    \centering
    \includegraphics[width=0.45\textwidth]{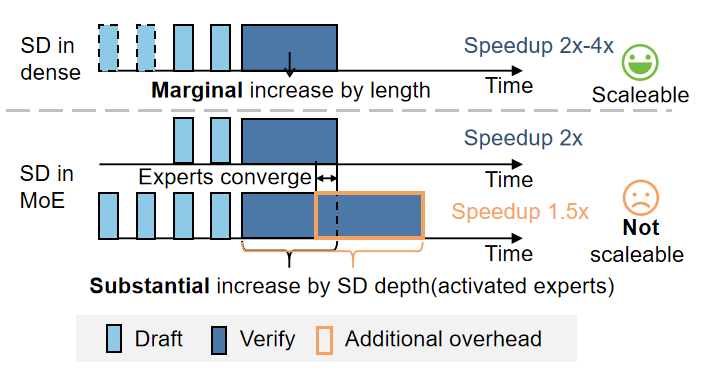} 
    \caption{Increasing the speculative depth yields an approximately linear benefit in dense models, but a markedly nonlinear trend in MoE models, as deeper speculation rapidly increases I/O demand under memory constraints.}
    \Description{Schematic timelines compare speculative decoding in dense and MoE models. Light blocks represent drafting, dark blocks represent verification, and an orange outline marks additional overhead. Increasing draft depth adds little verification time in the dense example, whereas the MoE example has a substantially longer verification interval and lower speedup because more experts are activated.}
    \label{fig:depth}
\end{figure}

Naive integration introduces severe system-level bottlenecks. While quantized SD provides a safe and high-fidelity lookahead mechanism, applying it directly to offloaded MoE inference introduces a severe system-level challenge.

\textit{VRAM contention limits expert caching.} The fundamental challenge begins at the memory level. MoE offloading inherently targets memory-constrained environments, a constraint that standard SD exacerbates by requiring the simultaneous maintenance of both a draft and a target model. Storing independent weights and maintaining redundant KV caches consumes precious GPU memory. Consequently, this redundancy leaves insufficient space for the crucial expert cache, directly degrading the system's baseline performance.

\begin{figure}[!htbp]
    \centering
    \includegraphics[width=\columnwidth]{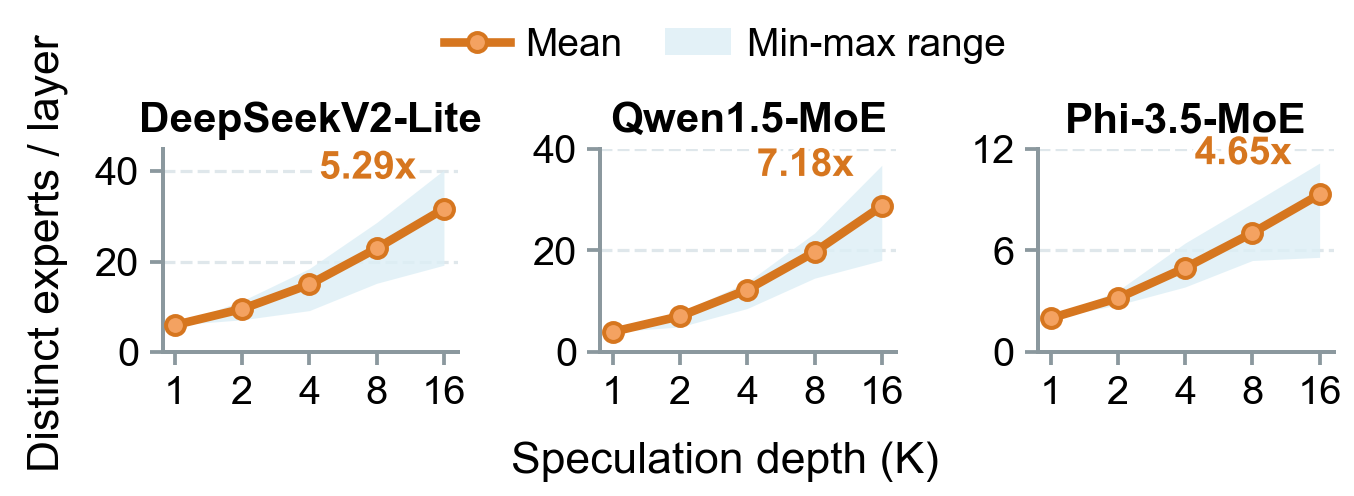}
    \caption{{Measured expert working-set growth with speculation depth.}}
    \Description{Three line charts plot distinct experts per layer against speculation depths of 1, 2, 4, 8, and 16 for DeepSeekV2-Lite, Qwen1.5-MoE, and Phi-3.5-MoE. Orange lines show the mean and shaded bands show the minimum-to-maximum range. Expert demand rises in all three models, with annotated growth factors of 5.29, 7.18, and 4.65, respectively.}
    \label{fig:expert-explosion-1}
\end{figure}

{{\textit{Deep speculation triggers I/O thrashing.}
Beyond static capacity, speculative decoding introduces severe dynamic I/O pressure. In memory-constrained MoE inference, increasing the speculation depth $K$ triggers an \textit{expert explosion} (Figure~\ref{fig:depth}), as predicting $K$ tokens requires the union of experts activated across all steps. Figure~\ref{fig:expert-explosion-1} quantifies this effect across three MoE models. For example, on DeepSeekV2-Lite scaling $K$ from $1$ to $16$ expands the per-layer expert demand from $6.0$ to $31.7$, which is a $5.3\times$ surge covering nearly half of the total $64$ experts. Once this enlarged working set exceeds the resident GPU cache, the required weights must be fetched synchronously over PCIe during target verification, causing heavy bus contention and critical-path I/O thrashing.
}}

\textit{Fragmented execution causes pipeline bubbles.} Even if the I/O bottleneck could be mitigated, the execution phase introduces its own inefficiencies. Traditional SD operates in a strictly serial manner, which fails to deeply overlap draft computation, PCIe transfers, and target verification, resulting in massive pipeline stalls. Furthermore, quantizing the draft model transforms already fine-grained MoE layers into tiny, fragmented INT4 kernels. Launching these fragmented kernels incurs disproportionately high overhead, leading to severe GPU underutilization during the critical draft phase.

\textbf{System-level co-design is imperative.} Ultimately, these cascading bottlenecks—from VRAM contention to I/O thrashing and fragmented execution—demonstrate that a piecemeal optimization is insufficient. Translating the theoretical multi-step lookahead of SD into actual end-to-end speedup in MoE offloading requires a holistic co-design of model state management, I/O scheduling, and kernel execution. This fundamental requirement motivates the design of \textsc{Specter}.

\begin{figure}[t]
    \centering
    \includegraphics[width=\columnwidth]{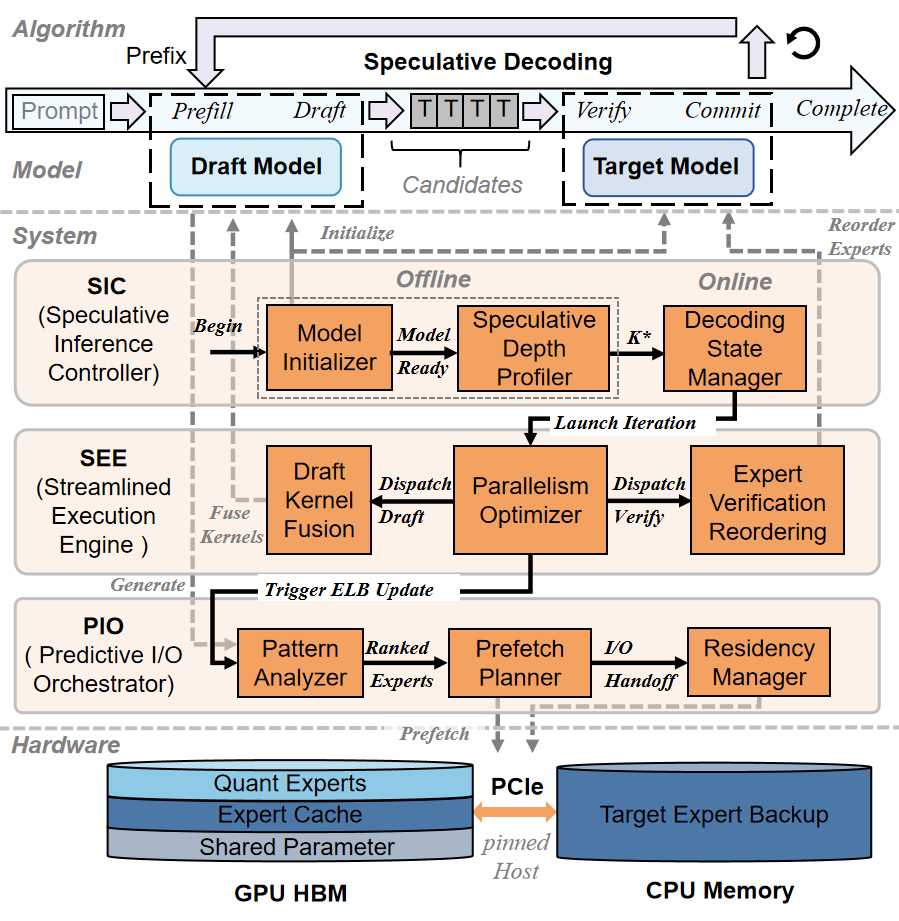}
    \caption{{The high-level architecture of \textsc{Specter}. The system integrates a Speculative Inference Controller, a Predictive I/O Orchestrator, and a Streamlined Execution Engine to transform I/O latency into productive computation through proactive data orchestration and adaptive control.}}
    \Description{A layered diagram connects the speculative decoding workflow to three system modules and the memory hierarchy. The controller initializes models, profiles speculative depth, and manages decoding state. The execution engine dispatches fused drafting and reordered expert verification. The I/O orchestrator analyzes routing patterns, plans prefetches, and manages residency. GPU memory holds quantized experts, an expert cache, and shared parameters, while target expert weights reside in CPU memory and transfer over PCIe.}
    \label{fig:overview}
\end{figure}

\section{\textsc{Specter}: System Architecture and Design}\label{sec:design}

\subsection{Design Principles and Architectural Overview}

\textsc{Specter} accelerates offloaded MoE inference by using
multi-step draft lookahead to overlap expert transfers with
computation, while parallel target verification improves
decoding efficiency. Three principles guide \textsc{Specter's} design.

First, \textbf{transforming latency through speculation}:
a lightweight quantized draft generates candidate tokens and
expert-activation predictions during the I/O latency window,
providing multi-step scheduling lookahead.

Second, \textbf{prediction-driven data orchestration}:
high-fidelity routing predictions from a 4-bit draft guide expert
prefetching, transfer timing, and verification order, using VRAM
as an actively managed staging area for locality and reuse.

Third, \textbf{efficiency through tightly integrated execution}:
model state sharing, hybrid precision, fused kernels, and
verification reordering jointly reduce draft, transfer, and
verification overhead under tight memory budgets.

As illustrated in Figure~\ref{fig:overview}, \textsc{Specter}
comprises three components: SIC coordinates hybrid-precision
model sharing and decoding-state management; PIO converts
multi-step lookahead into bandwidth-aware, three-phase PCIe
prefetching; and SEE reduces execution fragmentation through
kernel fusion, expert-wise reordering, and asynchronous overlap.

\subsection{Speculative Inference Controller}
\label{ssec:governor}

The Speculative Inference Controller serves as the control and organization layer of \textsc{Specter}. Rather than acting as a heavyweight online optimizer, it is responsible for organizing the speculative workflow, managing the draft and target model instances, and maintaining speculation-related runtime state. 
In particular, it coordinates interactions between the draft and target models.

This design reflects a practical consideration in offloaded MoE inference: the benefit of speculation does not come only from accurate prediction, but also from whether the system can organize model state and speculative information in a form that is directly usable by the expert scheduler and execution engine. The controller, therefore, provides a stable control interface that bridges speculative decoding, expert scheduling, and system execution.

The three SIC modules cover both the offline stage and online speculative inference: the Model Initializer constructs the shared draft--target instance, the Speculative Depth Profiler selects \(K\), and the Decoding State Manager runs each iteration and maintains committed token and KV states.

\begin{figure}[!htbp]
    \centering
    \includegraphics[width=\linewidth]{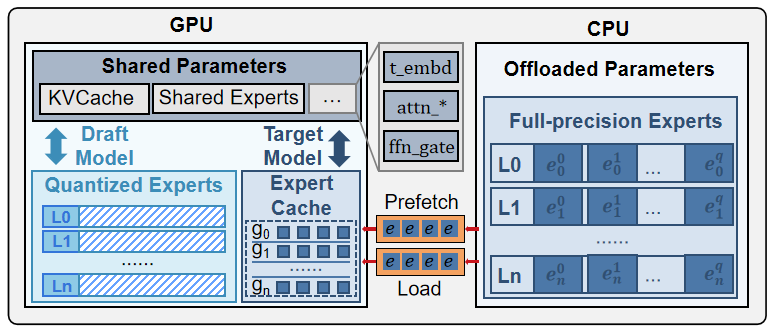}
    \caption{Overview of parameter distribution and data layout orchestration in \textsc{Specter}.}
    \Description{Diagram of parameter placement across GPU and CPU memory. The draft and target models share GPU-resident parameters and a KV cache. Quantized draft experts remain on the GPU, while full-precision target experts are prefetched or loaded from CPU memory into a GPU expert cache.}
    \label{fig:param_distribution}
\end{figure}

\subsubsection{Hybrid-Precision Model Initializer}

A key design choice in \textsc{Specter} is to share model state between the lightweight draft model and the full-precision target model whenever possible. Instead of maintaining two fully independent copies, the two models share the tensors associated with non-expert parameters, such as embeddings, attention layers, and normalization layers. Building on this shared-instance design, \textsc{Specter} adopts a carefully designed hybrid-precision layout for the MoE model in order to balance prediction fidelity, execution speed, and overall memory footprint.

\paragraph{Full-Precision Components (FP16).}

\begin{itemize}
\item \textbf{Non-Expert Parameters:} MoE routing is sensitive to quantization noise, which can alter expert selection~\cite{chen2025eac}. For example, the router logits directly determine which experts are selected for each token, and even small perturbations can change the top-k routing outcome. Since incorrect routing predictions directly degrade ELB quality, \textsc{Specter} keeps gating modules in FP16. The same applies to attention and other components outside the multilayer perceptron (MLP).
\item \textbf{Shared Experts:} In some MoE architectures, shared experts are activated for every token and remain resident in VRAM throughout inference. Since quantizing them brings limited offloading benefit but may affect the quality of the draft path, \textsc{Specter} also keeps these shared experts in FP16.
\end{itemize}

\paragraph{Quantized Components (INT4).}

All remaining expert MLP parameters are quantized to INT4. These parameters dominate the overall model size and are the primary target for reducing the memory footprint and accelerating efficient draft execution.

The resulting parameter placement strategy is illustrated in Figure~\ref{fig:param_distribution}.

\begin{figure}[t]
    \centering
    \includegraphics[width=0.49\textwidth]{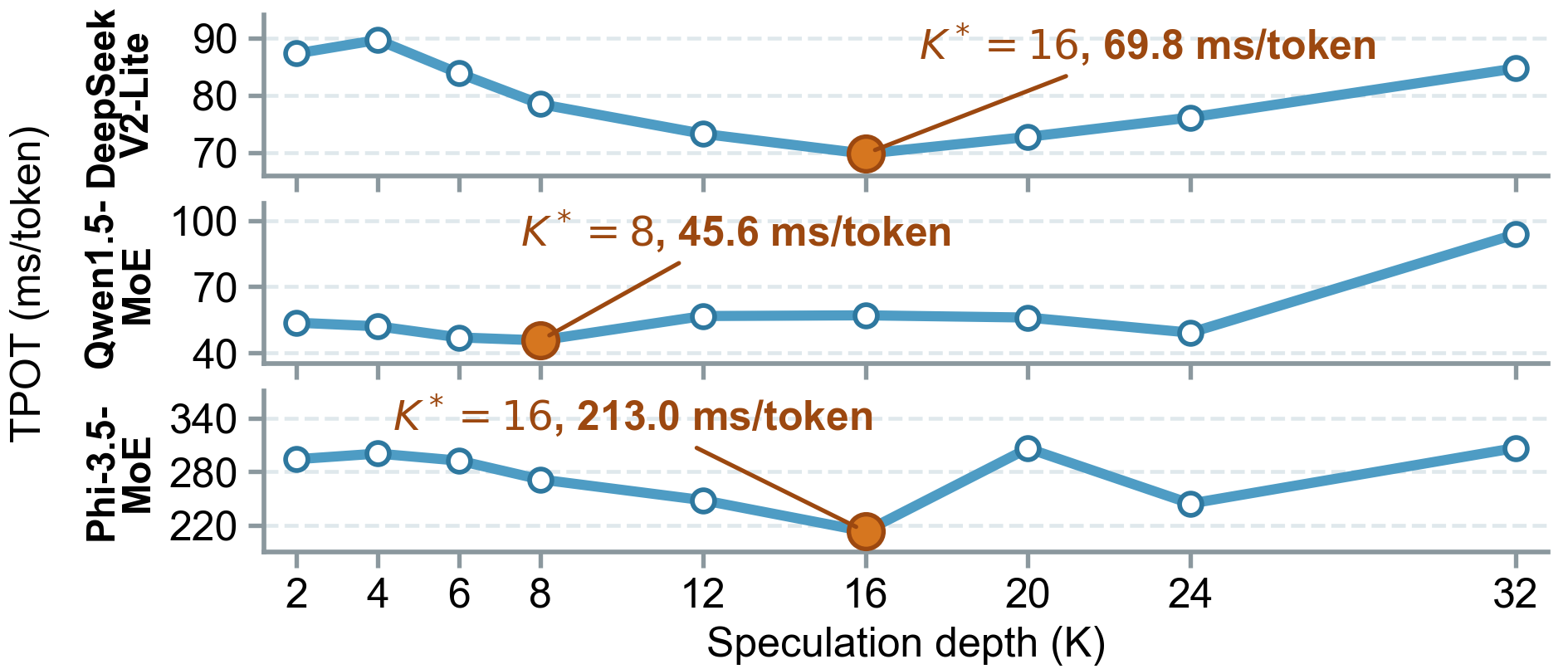}
    \caption{{Speculative depth profiling results for DeepSeekV2-Lite, Qwen1.5-MoE and Phi-3.5-MoE. The reported TPOT is averaged over prompts and repeated runs.}}
    \Description{Three line plots show TPOT against speculation depth for DeepSeekV2-Lite, Qwen1.5-MoE, and Phi-3.5-MoE. Highlighted minima occur at depths 16, 8, and 16, respectively. Increasing speculation depth beyond these minima does not consistently reduce TPOT.}
    \label{fig:offline-depth-profile}
\end{figure}

{{\subsubsection{Lightweight Speculative Depth Profiler.}
The speculation depth $K$ is a key system hyperparameter because it jointly determines the scheduler's lookahead, the distinct expert working set, and the costs of draft generation and target verification. In particular, the high acceptance rate of \textsc{Specter} (e.g., up to 99\% on DeepSeekV2-Lite) permits deeper speculation than the short windows commonly used by prior SD works~\cite{sadhukhan2025magicdec, su2023synergy}, making a fixed heuristic unreliable. 

We therefore select $K$ through lightweight offline profiling. We first estimate the mean token-acceptance rate $\bar{\alpha}$ from representative prompts. Following the analytical model commonly used by prior SD studies~\cite{leviathan2023fast}, the expected number of consecutively accepted draft tokens before the first rejection is
$\mathbb{E}[A]=\bar{\alpha}/(1-\bar{\alpha})$. Our measured
$\bar{\alpha}=96.63\%$ gives $\mathbb{E}[A]=28.7$, motivating the conservative upper bound $K_{\max}=32$. We sweep $\mathcal{K}=\{2,4,6,8,12,16,20,24,32\}$, evaluate 25 representative prompts three times, and select best $K$ by mean end-to-end TPOT:
\begin{equation}
K^{*}=\arg\min_{K\in\mathcal{K}}\operatorname{TPOT}(K).
\end{equation}

{As shown in Figure~\ref{fig:offline-depth-profile}, we select
$K=16$, $8$, and $16$ for DeepSeekV2-Lite, Qwen1.5-MoE, and
Phi-3.5-MoE, respectively.}}}

{{
\subsubsection{Target-Consistent Decoding State Management}
\label{ssec:kv_sharing}

\begin{figure}[t]
    \centering
    \includegraphics[width=0.45\textwidth]{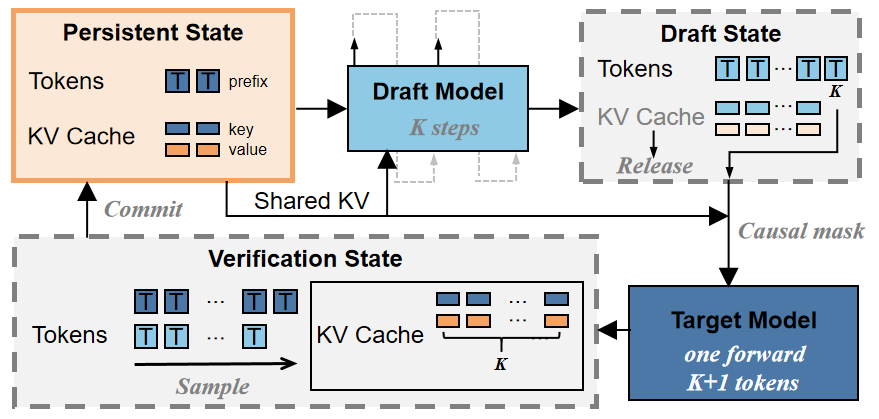}
    \caption{{State lifecycle across one speculative
    decoding iteration, including verifier-guided token commitment and
    shared-prefix KV management.}}
    \Description{State-transition diagram for one speculative decoding iteration. Persistent tokens and shared prefix KV states feed the draft and target models. Drafting creates temporary tokens and KV states; causally masked target verification produces verification states. Sampling commits accepted state to the persistent prefix, while temporary draft state is released.}
    \label{fig:state}
\end{figure}

The State Manager coordinates the authoritative token sequence and
shared KV states across speculative iterations. As illustrated in
Figure~\ref{fig:state}, each iteration begins from a target-consistent
prefix. The draft model extends this prefix with $K$ candidate tokens,
while the target model verifies them in a single causally masked forward
pass. Standard speculative sampling determines the accepted prefix and
produces either a corrected token after rejection or a bonus token when
all candidates are accepted, thereby preserving the target output
distribution. The State Manager commits these tokens to the
authoritative sequence and discards the rejected suffix.

The draft and target paths share the target-authoritative KV cache of
the committed prefix while maintaining separate temporary KV extensions
during speculation. The draft path records its speculative KV states in
a scratch extension, whereas target verification generates a
target-side extension for the candidate sequence. After verification,
only target-generated KV entries corresponding to accepted candidates
extend the shared cache. The draft extension and target-side entries for
the rejected suffix are discarded. The corrected or bonus token remains
the first uncached input of the next forward pass. This design avoids duplicating the persistent prefix KV cache while separating persistent state from temporary speculative extensions.
Each iteration therefore resumes from a target-authoritative sequence
and a valid KV boundary, reusing the committed prefix without retaining
draft-generated KV states.
}}

\subsection{Predictive I/O Orchestrator}
\label{ssec:scheduler}

{
The PIO makes the experts required by
target verification available in VRAM at the appropriate time, reducing
stalls caused by on-demand PCIe transfers. As illustrated in
Figure~\ref{fig:timeline_specter}, PIO coordinates three modules across
the draft and verification stages. During drafting, the Incremental
Routing Pattern Analyzer aggregates expert activations into layer-wise
routing metadata, from which the Layer-Aware Prefetch Planner generates
predictive host-to-device (H2D) transfers. Once target verification begins, the
Verification Expert Cache Residency Manager uses the target router's
authoritative routing decisions to coordinate a compute queue for cache-hit experts and an H2D queue for cache-miss experts, overlapping computation with the remaining H2D transfers.
}

{
\subsubsection{Incremental Routing Pattern Analyzer}
\label{sssec:pattern_analyzer}

The Incremental Routing Pattern Analyzer records the expert IDs produced
by the draft router, aggregates them into layer-wise routing metadata,
and uses this metadata to guide subsequent expert prefetching.
Conceptually, the ELB summarizes the top-$k$
routing decisions generated across $K$ draft tokens and $L$ MoE layers.
\textsc{Specter} implements it as a GPU-resident, layer-indexed
collection of activation counters. After observing the first $t$ draft
tokens, the counter for expert $e$ in layer $j$ is

\begin{equation}
\label{eq:elb_entry}
\mathrm{ELB}_{\mathrm{cnt}}^{(t)}[j,e]
=
\sum_{i=1}^{t}
\mathbb{I}\!\left[
e\in\mathrm{ELB}_{\mathrm{route}}[i,j]
\right],
\qquad t\leq K .
\end{equation}

Here, $\mathrm{ELB}_{\mathrm{route}}[i,j]$ denotes the logical
set of top-$k$ experts selected for draft token $i$ in layer $j$,
whereas $\mathrm{ELB}_{\mathrm{cnt}}^{(t)}[j,e]$ is the corresponding
cumulative activation counter. Only the counter representation is
materialized and maintained on the GPU.

At each prefetch checkpoint, the analyzer ranks experts in each layer by
their cumulative activation counts and selects the top-$M$ candidates,
where $M$ matches the expert-cache capacity assigned to that layer. The
selected layer-expert pairs are packed into a compact prefetch plan and
copied to the CPU runtime once per checkpoint. This checkpoint-level
aggregation avoids frequent per-token transfers and reduces
control-plane synchronization overhead.
After its final snapshot is consumed, the ELB is automatically cleared
at the end of the draft stage in each speculative iteration, preventing stale routing
evidence from carrying into the next draft window.

\begin{figure}[t]
    \centering
    \includegraphics[width=\columnwidth]{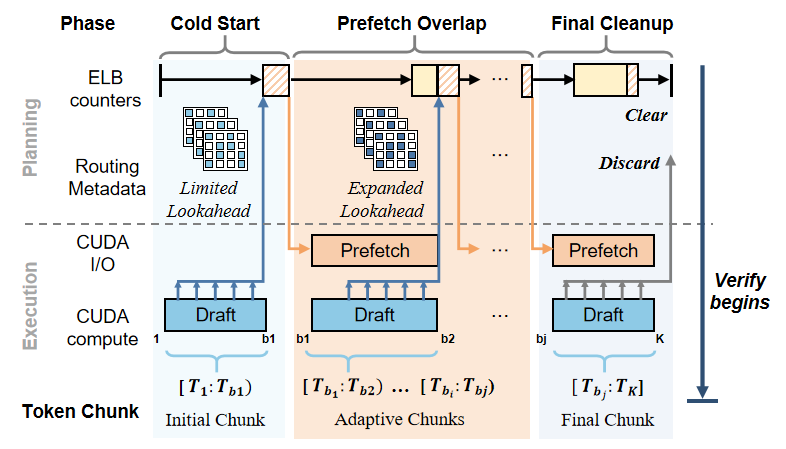}
    \caption{{Execution timeline of adaptive ELB-guided prefetching during drafting.}}
    \Description{Timeline divided into Cold Start, Prefetch Overlap, and Final Cleanup. ELB counters collect routing metadata during draft computation. Snapshots at chunk boundaries trigger prefetch transfers on a separate CUDA I/O stream, overlapping adaptive draft chunks. The final chunk discards new routing metadata and clears the ELB before verification begins.}
    \label{fig:timeline_specter}
\end{figure}

\subsubsection{Layer-Aware Prefetch Planner}
\label{sssec:prefetch_planner}
The Layer-Aware Prefetch Planner uses draft-time ELB lookahead to predict and prefetch experts likely to be required during target verification. As illustrated in Figure~\ref{fig:timeline_specter}, \textsc{Specter} organizes predictive I/O into three phases: \emph{Cold Start}, \emph{Prefetch Overlap}, and \emph{Final Cleanup}. A draft-token chunk is a contiguous subsequence of draft tokens executed as one compute-side overlap unit. Cold Start contains a single \emph{Initial Chunk}, Prefetch Overlap contains multiple \emph{Adaptive Chunks}, and Final Cleanup contains a single \emph{Final Chunk}. We use $b_i$ to denote the boundary between two adjacent chunks rather than a token index, and $T_{b_i:b_{i+1}}$ to denote the chunk enclosed by two consecutive boundaries. The initial boundary $b_1$ is predefined, while subsequent boundaries are determined online by the incremental prefetch workload.

At each boundary $b_i$, the Pattern Analyzer snapshots the accumulated ELB counters and generates a per-layer top-$M$ expert ranking. Based on the updated ranking, the planner identifies newly selected experts that are not already present in the cache. These incremental experts form the next H2D batch. The planner estimates the transfer time of this batch together with the draft-computation cost and selects $b_{i+1}$ such that the expected computation time of $T_{b_i:b_{i+1}}$ approximately matches the expected H2D time. The paired chunk and batch execute concurrently on the CUDA compute and I/O streams.

During Cold Start, the Pattern Analyzer collects routing metadata from the initial draft chunk. No predictive transfer is submitted before $b_1$ because the available ELB lookahead remains limited. The snapshot at $b_1$ generates the first incremental batch and determines the next chunk boundary. During Prefetch Overlap, this batch executes asynchronously while the corresponding draft chunk produces additional routing observations. At each successive boundary, the updated ELB snapshot generates another incremental batch, and the planner sizes the following chunk accordingly. This process continues until the next incremental batch is sufficient to fill the remaining expert-cache slots. Under the memory-constrained offloading configuration, this condition is reached before drafting completes.

The planner then enters Final Cleanup. The last incremental batch overlaps with the remaining draft computation and fills the expert cache. Within this phase, routing metadata produced by the final chunk is discarded and the ELB is cleared because the full cache cannot admit another predictive batch and target verification follows immediately. During verification, the Residency Manager manages expert I/O using the target router’s authoritative decisions.

\subsubsection{Verification Expert Cache Residency Manager}
\label{sssec:residency_manager}
The Verification Expert Cache Residency Manager coordinates cache
residency and expert I/O throughout target verification. Once
verification begins, the target router produces authoritative
expert-routing decisions for each layer. The manager classifies the
required experts according to their cache residency. \emph{Cache-hit
experts} are already resident in VRAM, either retained from previous
iterations or loaded by the Prefetch Planner, whereas \emph{cache-miss
experts} remain in host memory because of cache-capacity limits or
routing-prediction misses. The manager prioritizes cache-hit experts in
the compute queue while submitting cache-miss experts to the I/O queue,
overlapping expert computation with the remaining H2D transfers.
Each cache-hit expert remains protected while it is waiting in the
compute queue or processing its assigned token activations. Once all
tokens routed to that expert have completed computation, its slot is
unprotected and becomes eligible for replacement. When a cache-miss
expert finishes loading, it becomes resident and is appended to the
compute queue. This prevents loaded experts from being prematurely
replaced and avoids cache churn. The rolling load-compute process continues until every target-selected
expert in the current layer has processed its assigned token activations.
It then advances to the next layer, repeating the same residency and execution procedure until verification completes for this iteration.
}

\begin{figure}[t]
    \centering
    \begin{subfigure}[b]{0.23\textwidth}
        \centering
        \includegraphics[width=\linewidth]{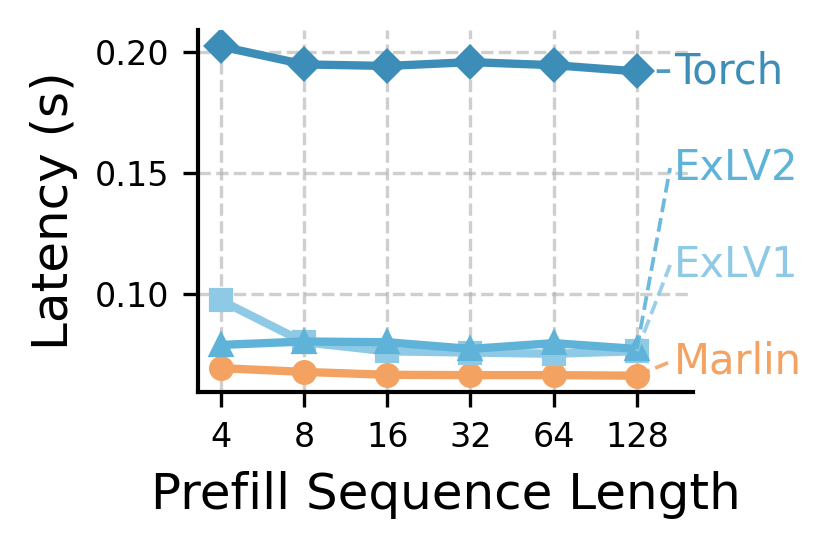}
        \caption{Prefill latency comparison}
        \label{fig:prefill}
    \end{subfigure}
    \hfill
    \begin{subfigure}[b]{0.23\textwidth}
        \centering
        \includegraphics[width=\linewidth]{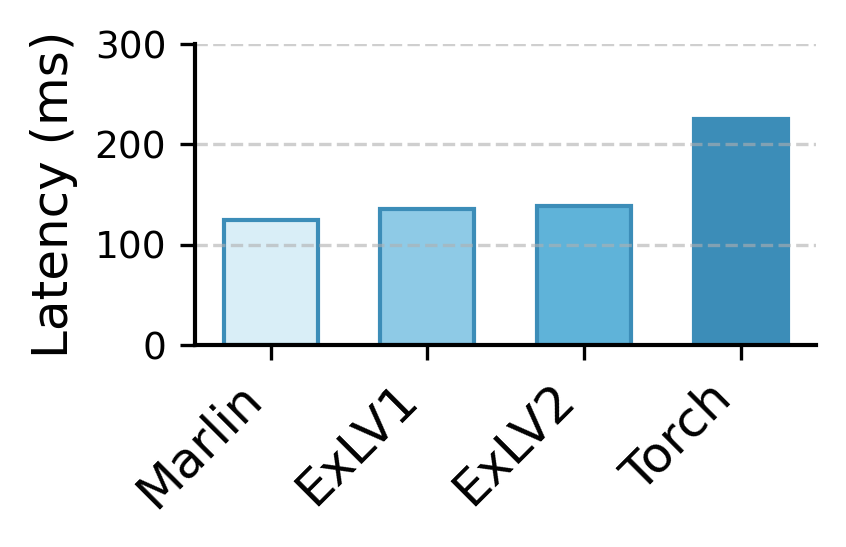}
        \caption{Decode latency comparison}
        \label{fig:decode}
    \end{subfigure}
    \caption{Performance comparison of different backends: (a) prefill stage latency, (b) decode stage latency. Lower values indicate better performance.}
    \Description{Two plots compare Marlin, ExLV1, ExLV2, and Torch. The prefill plot shows latency across sequence lengths from 4 to 128, and the decode plot shows one latency bar per backend. Marlin has the lowest latency and Torch the highest in both plots.}
    \label{fig:performance}
\end{figure}

\begin{figure}[t]
    \centering
    \includegraphics[width=0.48\textwidth]{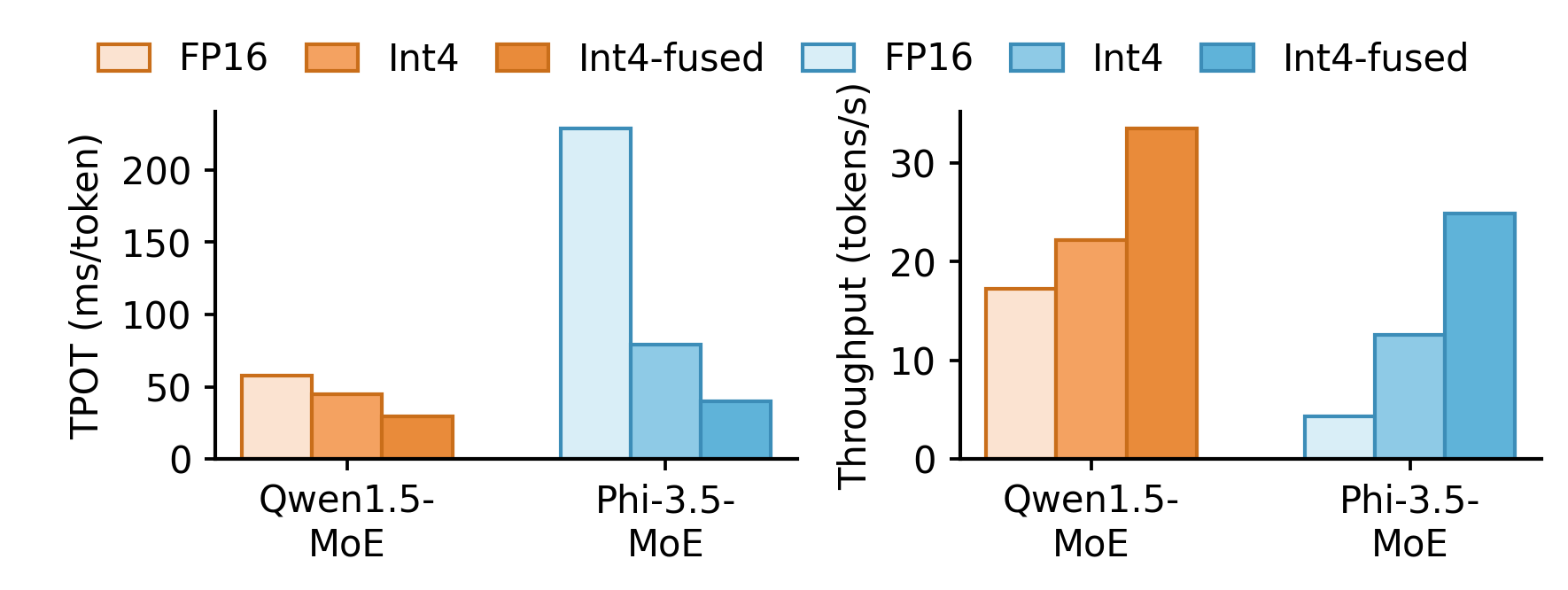}
    \caption{{Impact of quantization-aware fused expert kernels on throughput and latency in MoE models.}}
    \Description{Grouped bar charts compare FP16, INT4, and fused INT4 execution for Qwen1.5-MoE and Phi-3.5-MoE. For both models, INT4 lowers TPOT and raises throughput relative to FP16, and fused INT4 further lowers TPOT and raises throughput.}
    \label{fig:quant lantency throughput}
\end{figure}

\subsection{Streamlined Execution Engine}
\label{ssec:engine}

The end-to-end performance of \textsc{Specter} ultimately depends on the efficiency of its execution engine. Once the speculative depth and expert prefetch plan have been determined, the execution engine is responsible for three tightly coupled tasks: Draft Kernel Fusion accelerates proposal generation, Expert Verification Reordering restructures target-side MoE computation, and the Parallelism Optimizer coordinates computation and data movement to reduce pipeline stalls. In this way, the execution engine not only reduces kernel-level overhead, but also ensures that speculative prefetch can be effectively converted into practical end-to-end speedup. 

{
\subsubsection{Fine-Grained MoE Draft Kernel Fusion}

We adopt GPTQ~\cite{frantar2022gptq} for draft model quantization due to its favorable balance of implementation simplicity and accuracy retention. For execution backends, high-performance quantized engines such as Marlin~\cite{10.1145/3710848.3710871} achieve superior throughput during prefill and standard dense decoding, as illustrated in Figure~\ref{fig:performance}. However, Marlin is primarily optimized for large, regular general matrix‑matrix multiplications (GEMMs) (e.g., evaluating up to $72\mathrm{K}\times18\mathrm{K}$ matrices).
In contrast, modern fine-grained MoE architectures employ much narrower expert matrices (such as $2{,}048\times1{,}408$ or $4{,}096\times6{,}400$), where each activated expert receives only a small subset of the
$K$ speculative tokens. In this regime, frequent kernel-launch overheads and poor Tensor Core utilization across fragmented expert-local GEMMs counteract the benefits of low-bit arithmetic, occasionally rendering quantized execution slower than native FP16.

To overcome this efficiency gap, \textsc{Specter} introduces a fused multi-expert draft kernel. The kernel first gathers routed tokens by expert assignment and then schedules the disparate expert sub-computations into a single fused GPU launch. This unified dispatch amortizes launch latency, maximizes thread-block occupancy, and consolidates sparse token activations into dense computational tiles. 

{As shown in Figure~\ref{fig:quant lantency throughput},
both the unfused and fused INT4 variants use the Marlin backend.
This fusion improves draft decoding throughput on fine-grained
MoE models with minimal overhead.}

\par}

{{\subsubsection{Cache-Aware Expert Verification Reordering}
\label{sssec:reorder}

\begin{figure}[t]
    \centering
    \includegraphics[width=\columnwidth]{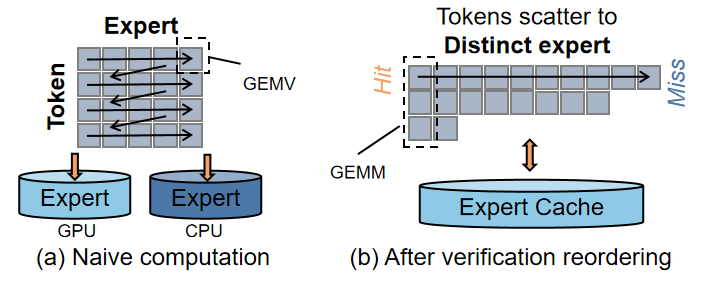}
    \caption{{Comparison of conventional token-wise verification and \textsc{Specter's} cache-aware expert-wise execution.}}
    \Description{Two diagrams compare token-wise and expert-wise verification. Naive computation traverses a token-by-expert grid and uses GEMV operations with experts in GPU or CPU memory. Reordering groups tokens by expert for GEMM computation and separates cache-hit and cache-miss experts around the expert cache.}
    \label{fig:reorder}
\end{figure}

In conventional autoregressive decoding with batch size one, the target model processes exactly one token per step, enforcing a strict chronological execution that precludes inter-token expert computation batching. Speculative decoding breaks this bottleneck by feeding a candidate sequence of $K$ tokens into the target model in a single forward pass. \textsc{Specter} leverages this verification lookahead window to fundamentally reorganize layer-wise MoE FFN computation. 

Figure~\ref{fig:reorder} contrasts conventional token-wise verification with \textsc{Specter's} cache-aware expert-wise execution. For each MoE layer, the target router first computes the authoritative routing decisions for all $K$ tokens. Rather than processing FFN of tokens in chronological order, \textsc{Specter} gathers tokens dispatched to identical experts and packs their activation states into contiguous sub‑batches. This spatial aggregation converts fragmented, memory-bound general matrix-vector multiplications (GEMVs) into dense, compute-bound GEMMs, maximizing the parallel compute capabilities of the GPU during target verification.

Furthermore, \textsc{Specter} integrates this reordering with a cache-aware execution schedule. It prioritizes the computation of cache-resident experts. Consider a speculative window where tokens route to both a resident expert A and a missing expert B. A naive sequential approach would trigger immediate I/O stalls and potential cache thrashing. Conversely, \textsc{Specter} aggregates and immediately processes the sub-batch for expert A. Concurrently, it dispatches asynchronous H2D transfers for expert B. This schedule effectively hides the PCIe I/O latency behind the aggregated GEMM computation of resident experts.

Upon completion of all expert sub-batches, \textsc{Specter} scatters the routing-weighted outputs. It accumulates them back into their original sequence positions. This transformation alters only the execution schedule of independent MoE FFN operations. It preserves the original target routing decisions. Attention mechanisms still evaluate using the standard causal mask. Finally, a layer-wise synchronization ensures all tokens complete before advancing. Consequently, this expert-wise reordering guarantees strict mathematical equivalence while substantially accelerating target verification.
}}

{
\subsubsection{Overlap-Oriented Parallelism Optimizer}

Beyond optimizing individual decoding phases and local components, \textsc{Specter} employs an Overlap-Oriented Parallelism Optimizer to reduce synchronization and data-movement stalls along its critical execution path. Efficient parallel execution follows two principles: preserving overlap through asynchronous execution with resource reuse, and eliminating avoidable synchronization from the critical path.

To realize these principles, the Overlap-Oriented Parallelism Optimizer coordinates module-specific optimizations across the execution pipeline. It first improves Draft Kernel Fusion by preallocating and reusing intermediate activation buffers and tuning the shared-memory allocation of expert MLP kernels. Within the Pattern Analyzer, it incrementally accumulates ELB counters on the GPU and transfers only a compact expert plan through a batched device-to-host (D2H) operation at each prefetch point, avoiding frequent host synchronization after individual routing updates. Finally, it coordinates the Prefetch Planner and Residency Manager using dedicated CUDA streams to separate computation from data movement and CUDA events to enforce expert-readiness and safe cache-slot-reuse dependencies. It further combines pinned host memory with non-blocking copies to support efficient asynchronous H2D transfers. 
\par}

\begin{table}[!htbp]
\centering
\caption{Configuration of evaluated models.}
\label{tab:moe_models}

\begingroup
\setlength{\tabcolsep}{4pt}
\resizebox{\columnwidth}{!}{%
\begin{tabular}{lccc}
\toprule
& Phi-3.5-MoE~\cite{abdin2024phi3technicalreporthighly} & Qwen1.5-MoE~\cite{qwen_moe} & DeepSeekV2-Lite~\cite{deepseekv2} \\
\midrule
Total Param.        & 41.9B             & 14.3B & 15.7B \\
Activated Param.    & 6.6B              & 2.7B  & 2.4B  \\
Total Weight        & 78GB              & 29GB  & 34GB  \\
\midrule
MoE Layer Number    & 32                & 24    & 26    \\
First Dense Layer   & 0                 & 0     & 1     \\
Experts (Per Layer) & 16                 & 60    & 64    \\
Top-$k$             & 2                 & 4     & 6     \\
Shared Experts      & 0                 & 1(4x Dim + gate) & 1(2x Dim) \\
Hidden Size         & 4096              & 2048  & 2048  \\
MoE Inter. Size     & 6400              & 1408  & 1408  \\
\bottomrule
\end{tabular}%
}
\endgroup
\end{table}

\begin{figure}[!htbp]
    \centering
    \includegraphics[width=\columnwidth]{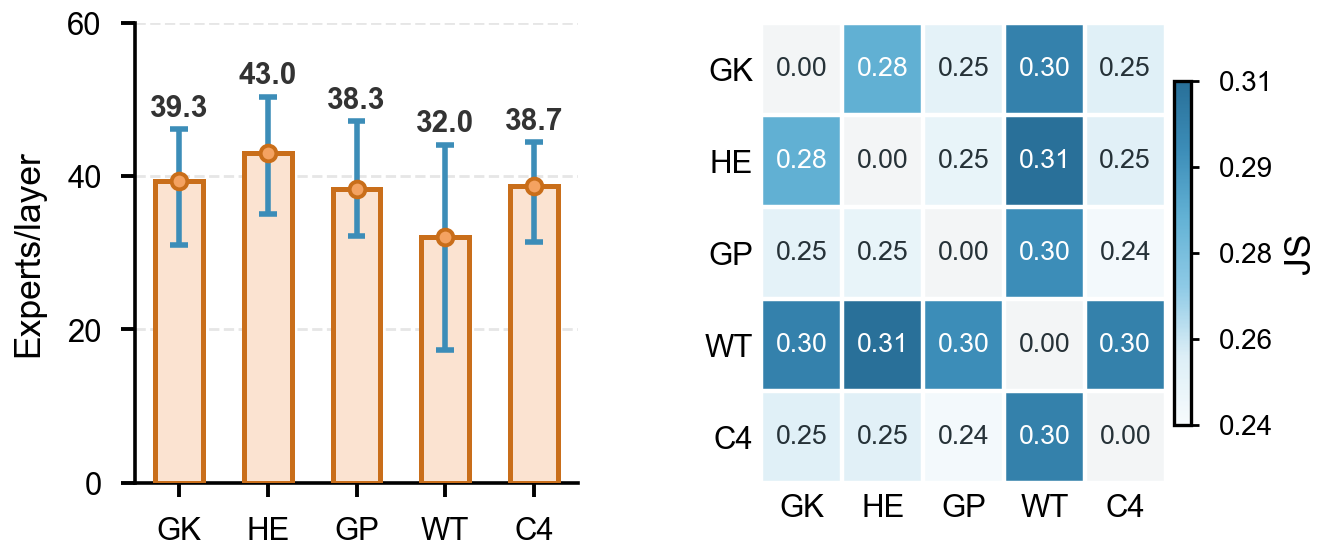}
    \caption{{Dataset-induced expert-routing diversity on
    DeepSeekV2-Lite with $K=16$. Left: mean distinct experts per layer of each dataset, with 10th--90th percentile ranges. Right: pairwise Jensen--Shannon distances between normalized routing distributions.}}
    \Description{Two panels compare five datasets. The bar chart shows distinct experts per layer with percentile intervals: HumanEval has the largest mean and WikiText-2 the smallest. The heat map shows zero self-distances and nonzero distances between different datasets, with WikiText-2 generally the most distinct.}
    \label{fig:dataset-routing-diversity}
\end{figure}

\section{Evaluation}

\subsection{Experimental Setup}

\subsubsection{Hardware and Software Platform}

All experiments are conducted on a workstation with one NVIDIA A100 GPU with 40 GB of HBM memory and a 24-core Intel Xeon Silver 4310 CPU with 256 GB RAM. The GPU is connected to the CPU via a PCIe 4.0 x16 interface, providing a theoretical bandwidth of 31.5 GB/s for H2D data transfers. Additionally, we evaluate \textsc{Specter} under several representative GPU memory budgets in order to systematically study its sensitivity to available device memory under offloading.

{{Our system is implemented on Hugging Face Transformers~\cite{wolf-etal-2020-transformers}.
The Transformers implementation uses GPTQ~\cite{frantar2022gptq} for
draft-model quantization. We use Hugging Face Transformers 4.53.3, PyTorch 2.7.0, CUDA Toolkit 12.8, GPTQModel~\cite{qubitium2024gptqmodel} at commit \texttt{037c5c0}, and its integrated Marlin.}}

\begin{figure*}[ht]
    \centering
    \includegraphics[width=\textwidth]{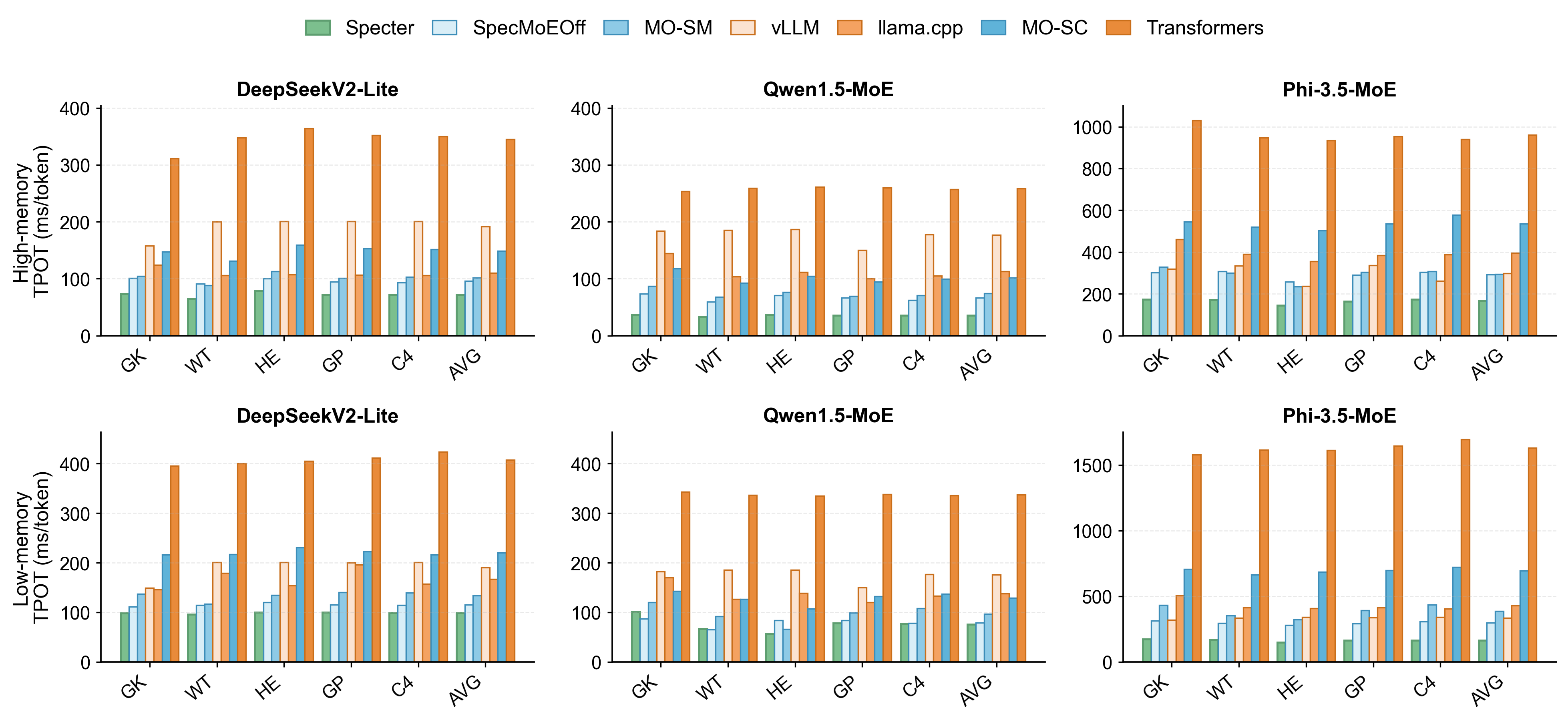}
    \caption{{End-to-end inference performance of \textsc{Specter} and other approaches on varying datasets, memory settings and models. The figures in the upper row represent experiments conducted under higher GPU memory conditions, while the figures in the bottom row represent experiments performed under lower GPU memory conditions. Under shared-weight accounting, the reference high/low memory budgets are 24/18\,GB, 20/16\,GB, and 40/32\,GB for DeepSeekV2-Lite, Qwen1.5-MoE, and Phi-3.5-MoE, respectively.}}
    
    \Description{Six grouped bar charts show TPOT in milliseconds per token for seven systems across five datasets and their average. Columns represent DeepSeekV2-Lite, Qwen1.5-MoE, and Phi-3.5-MoE; rows represent high and low memory budgets. Specter has the lowest average TPOT in each panel, while Transformers has the highest.}
    \label{fig:end to end}
\end{figure*}

\subsubsection{Models and Workloads}
We select three representative mainstream MoE models in Table \ref{tab:moe_models} that exhibit substantial differences in overall scale, architectural design, and MoE layer implementation.
We use the official FP16 weights for the target and a GPTQ-quantized draft. Specifically, all expert-module linear layers use symmetric INT4 quantization with a group size of 128.

To evaluate performance across diverse domains, we use five standard benchmark datasets:  C4~\cite{JMLR:v21:20-074} (web-crawled corpus), WikiText-2-v1~\cite{merity2016wikitext} (factual language modeling), HumanEval~\cite{chen2021evaluating} (code generation), GSM8K~\cite{cobbe2021gsm8k} (mathematical reasoning), and GPQA~\cite{rein2024gpqa} (expert-level multidisciplinary questions). In the following experiments and figures, we use the abbreviations GK (GSM8K), WT (WikiText-2), HE (HumanEval), GP (GPQA), C4, and AVG (average) to refer to the respective datasets for brevity.

{{Figure~\ref{fig:dataset-routing-diversity} further verifies workload
diversity at the routing level. The left panel shows that different
datasets produce different per-window expert working sets, while the
right panel reveals clear pairwise differences between their routing
distributions. These results confirm that the selected prompts exercise
diverse expert-demand patterns, rather than a single favorable routing
behavior, supporting the robustness of our evaluation across domains.}}

\subsubsection{Baselines}

We evaluate \textsc{Specter} against six baselines spanning both mainstream serving frameworks and research-oriented offloading designs. The first group consists of three widely used inference frameworks: vLLM~\cite{kwon2023efficient}, Transformers~\cite{wolf-etal-2020-transformers}, and llama.cpp~\cite{llamacpp2025}. These baselines reflect practical deployment stacks with different implementation philosophies and provide a broad view of how \textsc{Specter} compares against commonly used runtime systems. Concretely, their default offloading behaviors differ substantially. vLLM exposes CPU offloading as an on-demand weight offloading mechanism. Transformers represents a more generic coarse-grained offloading baseline, in which CPU-resident layers are loaded on demand during execution. In contrast, llama.cpp follows a CPU--GPU collaborative execution style, where a configurable number of layers are offloaded to the GPU while the remaining layers stay on the CPU, making it a practical hybrid baseline for edge local deployment.

The second group consists of three research baselines tailored to MoE offloading: Mixtral-Offloading-SC (MO-SC), Mixtral-Offloading-SM (MO-SM)~\cite{eliseev2023fast}, and SpecMoEOff~\cite{wang2025accelerating}. Mixtral-Offloading represents a specialized MoE offloading design based on expert-level, on-demand swapping between GPU and host memory, and our cache mechanism is also implemented on top of its cache-management foundation. We include both the same-cache (SC) and same-memory (SM) variants to cover two relevant comparison settings: SC reflects a more favorable cache regime, while SM provides a fairer comparison under same single-GPU memory budget.

\subsection{End-to-End Performance}
\subsubsection{Single-Request Performance}

Figure~\ref{fig:end to end} reports single-request end-to-end inference performance, measured as TPOT, for \textsc{Specter} and all baselines. We evaluate three representative MoE models on five datasets under both high-memory and low-memory settings. Across the evaluated models and memory configurations, \textsc{Specter} achieves the lowest average TPOT. We use SpecMoEOff as the primary MoE speculative-offloading baseline and \textsc{Specter} achieves an average $1.50\times$ TPOT speedup and up to $1.88\times$.

The advantage is especially clear in the high-memory setting, where reduced memory pressure makes the execution efficiency of each system easier to expose. On Phi-3.5-MoE, \textsc{Specter} achieves an average TPOT of 163.3\,ms, compared with 292.1\,ms for SpecMoEOff, 295.1\,ms for Mixtral-Offloading-SM, 297.5\,ms for vLLM, and 960.6\,ms for Transformers, corresponding to up to a 5.9$\times$ speedup over the generic Transformers baseline. On Qwen1.5-MoE, \textsc{Specter} reaches 35.3\,ms/token, substantially outperforming SpecMoEOff (66.4\,ms), Mixtral-Offloading-SM (74.2\,ms), vLLM (176.6\,ms), and Transformers (258.2\,ms). Even on DeepSeekV2-Lite, where the gap among stronger baselines is smaller, \textsc{Specter} still delivers the best overall performance, reducing average TPOT from 344.9\,ms for Transformers to 72.2\,ms.

The same trend persists in the low-memory setting. Although tighter memory budgets amplify the cost of expert movement and narrow the gap between some baselines, \textsc{Specter} achieves the lowest TPOT on most datasets and in the per-model averages. In particular, it continues to outperform both speculative and non-speculative alternatives. This result suggests that the benefit of \textsc{Specter} does not depend on a single favorable memory regime; rather, it continues to translate speculative lookahead into practical gains even when expert residency becomes more constrained.

\begin{figure}[!htbp]
    \centering
    \includegraphics[width=0.49\textwidth]
    {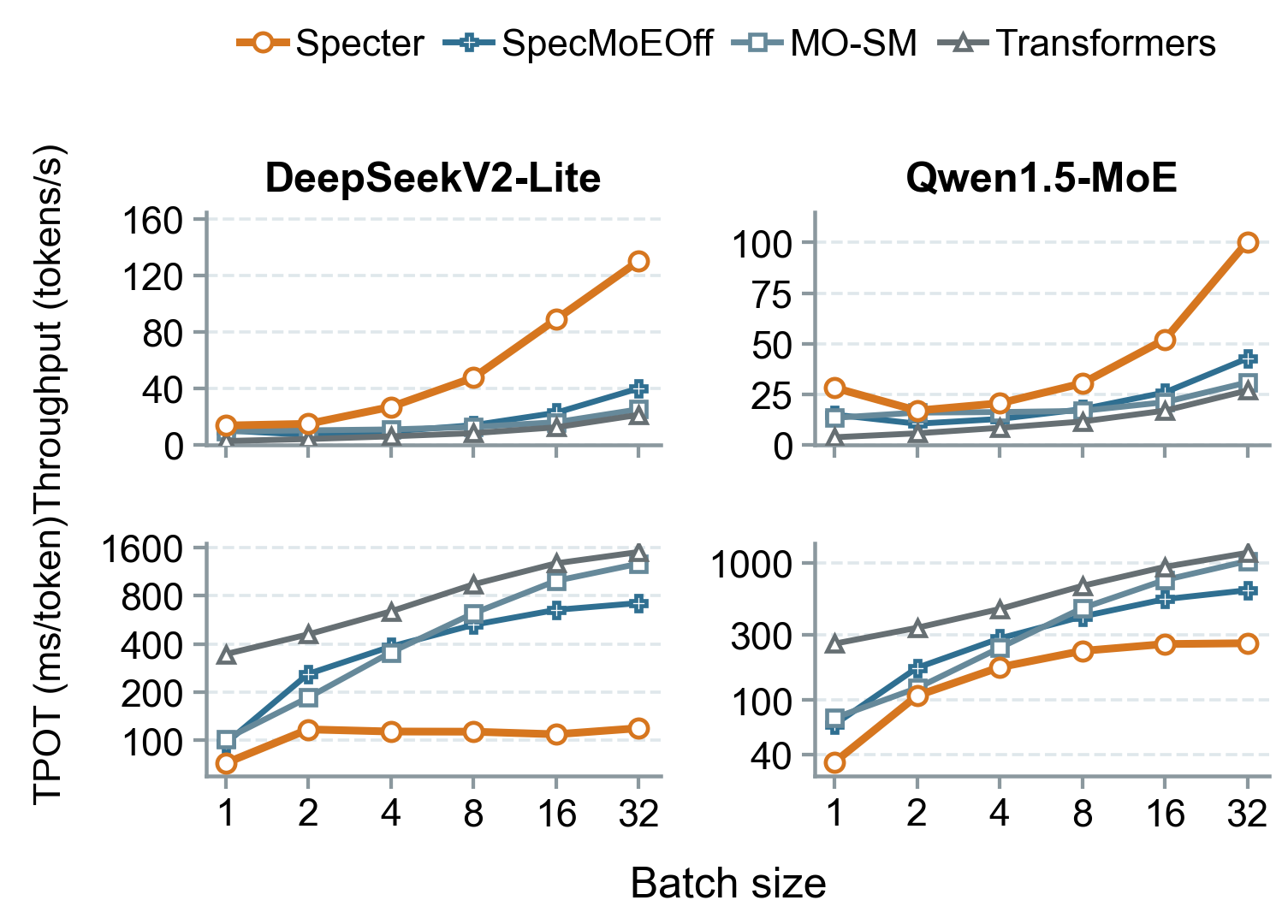}
    \caption{{TPOT and throughput scaling with batch size on DeepSeekV2-Lite and Qwen1.5-MoE.}}
    \Description{Four line charts compare Specter, SpecMoEOff, MO-SM, and Transformers as batch size increases from 1 to 32. Columns represent DeepSeekV2-Lite and Qwen1.5-MoE; the upper row shows throughput and the lower row shows TPOT on logarithmic vertical scales. Specter has the highest throughput and lowest TPOT at every displayed batch size.}
    \label{fig:batch-scalability}
\end{figure}

\begin{table*}[t]
\centering
\caption{{Absolute performance across workloads at $B=32$.}}
\label{tab:batch-scalability}

\footnotesize
\renewcommand{\arraystretch}{1.00}
\setlength{\tabcolsep}{2.5pt}

\begin{tabular*}{0.92\textwidth}{
    @{\extracolsep{\fill}} l *{12}{c} @{}
}
\toprule
& \multicolumn{6}{c}{DeepSeekV2-Lite}
& \multicolumn{6}{c}{Qwen1.5-MoE} \\
\cmidrule(lr){2-7}\cmidrule(lr){8-13}
Method
& GK & WT & HE & GP & C4 & AVG
& GK & WT & HE & GP & C4 & AVG \\
\midrule
\multicolumn{13}{c}{\textit{TPOT (ms/token) $\downarrow$}} \\
\midrule
Transformers
& 1504.7 & 1451.2 & 1489.6 & 1485.9 & 1536.7 & 1493.6
& 1171.0 & 1192.2 & 1129.0 & 1195.6 & 1249.2 & 1187.4 \\
MO-SM
& 1267.3 & 1206.5 & 1254.9 & 1268.6 & 1317.0 & 1262.9
& 1059.5 & 1020.6 & 982.8 & 1038.5 & 1067.1 & 1033.7 \\
SpecMoEOff
& 707.6 & 722.0 & 705.7 & 705.7 & 749.0 & 718.0
& 686.2 & 600.2 & 593.2 & 641.1 & 648.1 & 633.8 \\
\textbf{Specter}
& \textbf{125.9} & \textbf{125.0} & \textbf{100.2}
& \textbf{122.9} & \textbf{122.1} & \textbf{119.2}
& \textbf{286.6} & \textbf{250.0} & \textbf{241.6}
& \textbf{256.3} & \textbf{267.1} & \textbf{260.3} \\
\midrule
\multicolumn{13}{c}{\textit{Throughput (tokens/s) $\uparrow$}} \\
\midrule
Transformers
& 21.2 & 22.0 & 21.5 & 21.5 & 20.8 & 21.4
& 27.3 & 26.8 & 28.3 & 26.7 & 25.6 & 26.9 \\
MO-SM
& 25.2 & 26.5 & 25.5 & 25.2 & 24.3 & 25.3
& 30.2 & 31.3 & 32.5 & 30.8 & 30.0 & 30.9 \\
SpecMoEOff
& 41.1 & 38.6 & 40.7 & 40.8 & 39.5 & 40.2
& 40.8 & 42.5 & 47.0 & 41.9 & 41.7 & 42.8 \\
\textbf{Specter}
& \textbf{160.2} & \textbf{125.6} & \textbf{127.0}
& \textbf{107.3} & \textbf{131.4} & \textbf{130.3}
& \textbf{90.6} & \textbf{97.5} & \textbf{110.6}
& \textbf{106.2} & \textbf{96.7} & \textbf{100.3} \\
\bottomrule
\end{tabular*}
\end{table*}

\begin{figure}[!htbp]
    \centering
    \includegraphics[width=\columnwidth]{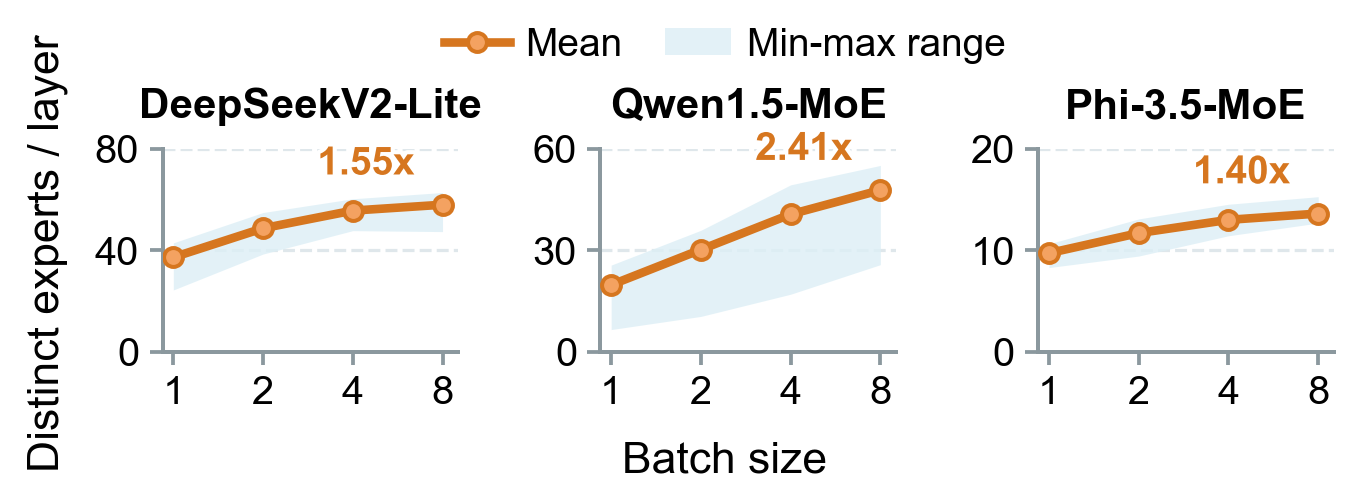}
    \caption{{Expert working-set growth with batch size.}}
    \Description{Three line charts show mean distinct experts per layer and shaded minimum-to-maximum ranges for batch sizes 1, 2, 4, and 8. Expert counts increase for DeepSeekV2-Lite, Qwen1.5-MoE, and Phi-3.5-MoE, with Qwen1.5-MoE exhibiting the largest relative increase.}
    \label{fig:expert-explosion-2}
\end{figure}

\subsubsection{Batch-Size Evaluation}

We extend \textsc{Specter} to support batched execution, while
maintaining independent sequence and KV states, acceptance decisions,
and rollback boundaries for each request.
{MO-SM loads experts across the full batch,
SpecMoEOff uses two microbatches sharing an expert cache, and
Transformers uses native batched execution.}
We evaluate DeepSeekV2-Lite and Qwen1.5-MoE on five datasets with
{$B=\{1,2,4,8,16,32\}$ under the same high-memory
settings as in Section~4.2.1}, using 16-token prefixes and generating
128 tokens per request.
Figure~\ref{fig:batch-scalability} reports
{TPOT and aggregate throughput averaged across
the five datasets}, while Table~\ref{tab:batch-scalability} provides
the per-dataset results at $B=32$.
{Mean TPOT is the average per-request inter-token latency, following the standard definition, whereas throughput divides total generated tokens by the batch wall time until the last request completes its generation.}

{From $B=1$ to $B=32$, \textsc{Specter's} TPOT
reaches $1.65\times$ and $7.37\times$ its single-request values on
DeepSeekV2-Lite and Qwen1.5-MoE, respectively, while throughput
improves by $9.41\times$ and $3.54\times$.
The largest relative TPOT increase occurs from $B=1$ to $B=2$:
$1.62\times$ on DeepSeekV2-Lite and $3.05\times$ on Qwen1.5-MoE.
Against SpecMoEOff, the maximum gains are a $6.02\times$ mean-TPOT
speedup at $B=32$ and $3.89\times$ the mean throughput at $B=16$,
both on DeepSeekV2-Lite.}

The TPOT increase is consistent with the batch-induced
expert-working-set growth shown in Figure~\ref{fig:expert-explosion-2}:
larger batches activate more distinct experts during verification, increasing target computation and H2D pressure.
{MO-SM and Transformers advance one token per
request at each batched decoding step, whereas \textsc{Specter}
verifies multiple candidate tokens per request together.
SpecMoEOff also supports speculative execution, but incurs
additional expert transfers between its microbatches as working
sets grow under limited cache capacity.}
{The $B=1$-to-$2$ transition introduces batch-specific
handling of per-request progress and KV states, which weigh
more heavily on Qwen1.5-MoE's shorter single-request execution.
Its expert working set also expands more in relative terms
($1.52\times$ versus $1.31\times$ for DeepSeekV2-Lite).}

\subsection{Memory Saving Analysis}

\begin{table}[!htbp]
\centering
\caption{Comparative GPU memory footprints (GB) of full-precision and 4-bit quantized Qwen1.5-MoE with a 12K-token context.}
\label{tab:shared_vram_footprint}
\resizebox{\columnwidth}{!}{
\begin{tabular*}{0.5\textwidth}{c|cccc}
\toprule
{\textbf{Scenario}} & {\textbf{Draft}} & {\textbf{Target}} & {\textbf{Shared}} & {\textbf{Total}} \\
\midrule
Independent    & 6.13 & 7.27 & {---} & 13.40 \\
+ Shared Experts     & 4.58 & 5.72 & 1.55  & 11.85 \\
+ Other Params       & 2.66 & 3.80 & 3.47  & 9.93  \\
+ KV Cache           & 0.41 & 1.55 & 5.72  & \textbf{7.68} \textsuperscript{$\downarrow$43\%} \\
\bottomrule
\end{tabular*}
}
\end{table}

Architectural alignment between the quantized draft and full-precision target enables extensive sharing of model parameters and runtime state, significantly reducing the GPU memory footprint of speculative decoding.

Table~\ref{tab:shared_vram_footprint} quantifies these savings for the Qwen1.5-MoE model in a 12K context setting. Under the evaluated  configuration, GPU memory stores all INT4 draft experts and 4 FP16 target experts per layer. The analysis shows a progressive reduction in memory usage as more components are shared. By co-locating shared experts, non-expert layers (e.g., embeddings and routers), and finally the token-level KV cache, the total memory requirement is incrementally lowered. A key enabler for the final step is our state synchronization mechanism, which allows only target-validated KV entries to extend the shared prefix cache after each verification step. This comprehensive sharing strategy ultimately culminates in a total memory footprint of 7.68 GB, achieving a 43\% reduction compared to a naive implementation where both models operate fully independently.

\subsection{Impact of Prefetch Strategy} 

\begin{figure}[!htbp]
    \centering
    \includegraphics[width=\columnwidth]{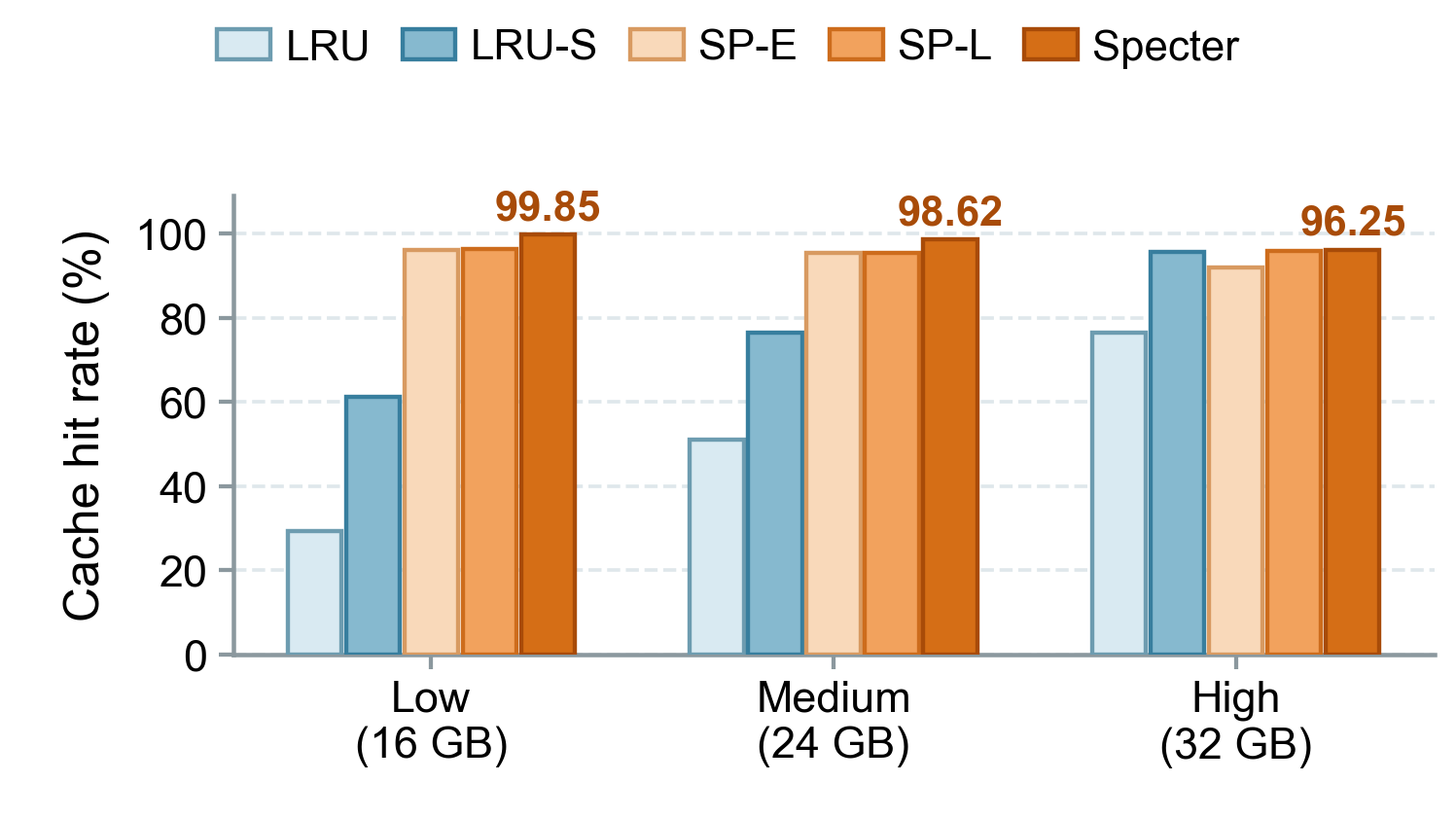}
    \caption{Cache hit rates of different policies on DeepSeekV2-Lite under three GPU-memory budgets.}
    \Description{Grouped bars compare cache hit percentages for LRU, LRU-S, SP-E, SP-L, and Specter at memory budgets of 16, 24, and 32 GB. Specter remains close to 100 percent and has the highest hit rate in each group. LRU and LRU-S improve as the memory budget increases, with LRU-S approaching the prefetch policies at 32 GB.}
    \label{fig:cache_hit_rate}
\end{figure}

{{We compare five cache-management policies in
Figure~\ref{fig:cache_hit_rate}. \emph{LRU} is purely reactive: an
expert is loaded only after being selected by the target router, and the
least recently used expert in the same layer is evicted when necessary.
\emph{LRU-S} applies the same replacement policy with additional cache
slots under the same total GPU-memory budget, isolating the benefit of
increased cache capacity. \emph{SP-E} and \emph{SP-L} use the same ELB
predictions and base cache capacity as \textsc{Specter}, but each prefetches once at its respective draft checkpoint.

We conduct this study on DeepSeekV2-Lite under 16\,GB, 24\,GB, and
32\,GB memory budgets. Cache hit rate is defined as the fraction of
target-routed experts already available in GPU cache before on-demand
loading, averaged across verification steps. Both single-prefetch
policies substantially outperform reactive LRU, demonstrating the value
of draft-derived lookahead. By repeatedly refining the ELB and issuing incremental prefetch batches,
\textsc{Specter} achieves the highest hit rate under every budget:
99.85\%, 98.62\%, and 96.25\%, respectively. LRU-S becomes competitive
at 32\,GB because its larger cache retains more of the expert working
set, but this improvement comes from additional cache capacity rather
than predictive data movement.}}

\subsection{System Component Analysis}

\begin{table}[!htbp]
\centering
\caption{{Per-iteration latency breakdown across models and cumulative component ablation on DeepSeekV2-Lite.}}
\label{tab:latency_breakdown_revision}

\textit{(a) Per-iteration latency breakdown}\par\vspace{2pt}
\resizebox{\columnwidth}{!}{
\begin{tabular}{lccc}
\toprule
Metric & DeepSeekV2-Lite & Qwen1.5-MoE & Phi-3.5-MoE \\
\midrule
Draft             & 674.0 & 223.3 & 952.5  \\
Verification      & 136.1 & 88.2  & 82.2   \\
Total H2D Load    & 243.4 & 89.7  & 2089.2 \\
Exposed H2D Stall & 108.3 & 35.4  & 1621.3 \\
\textbf{Hidden H2D}        & \textbf{135.1} & \textbf{54.3} & \textbf{467.9} \\
\textbf{Hidden Ratio}      & \textbf{55.5\%} & \textbf{60.5\%} & \textbf{22.4\%} \\
\bottomrule
\end{tabular}}

\vspace{4pt}
\textit{(b) Cumulative component ablation}\par\vspace{2pt}
\resizebox{\columnwidth}{!}{
\begin{tabular}{lcc}
\toprule
Configuration & TPOT (ms/token) $\downarrow$ & Slowdown \\
\midrule
\textbf{Full \textsc{Specter}} & \textbf{69.2} & \textbf{1.00$\times$} \\
w/o PIO                       & 83.5  & 1.21$\times$ \\
w/o PIO + SIC                 & 94.4  & 1.36$\times$ \\
w/o PIO + SIC + SEE           & 130.4 & 1.88$\times$ \\
\bottomrule
\end{tabular}}
\end{table}

{{
Table~\ref{tab:latency_breakdown_revision}(a) directly quantifies the
latency hidden by \textsc{Specter}. It overlaps 55.5\%, 60.5\%, and 22.4\% of their total loading time on DeepSeekV2-Lite,
Qwen1.5-MoE, and Phi-3.5-MoE, respectively. These transfer intervals overlap
draft generation and target verification and therefore should not be
summed as sequential latency components. Although Phi-3.5-MoE hides the
largest absolute amount, its substantially larger transfer demand leaves
more H2D latency exposed.

Table~\ref{tab:latency_breakdown_revision}(b) attributes the 
improvement through cumulative module removal. Disabling PIO increases
TPOT from 69.2 to 83.5\,ms/token; additionally removing SIC raises it to
94.4\,ms/token; removing SEE as well reaches 130.4\,ms/token. 
}}

\begin{table}[!htbp]
\centering
\caption{{Robustness to expert-prediction errors and natural
variations. TPOT is reported in ms/token.}}
\label{tab:robustness}

\begingroup
\small
\renewcommand{\arraystretch}{1.04}
\setlength{\tabcolsep}{1.2pt}

\begin{tabular*}{\columnwidth}{
@{\extracolsep{\fill}} l c c c c @{}
}
\toprule
\multicolumn{5}{c}{(a) Controlled expert-prediction errors} \\
\midrule
\textbf{Error Rate} & \textbf{0\%} & \textbf{25\%}
                    & \textbf{50\%} & \textbf{75\%} \\
\midrule
Prefetch Acc. & 1.000 & 0.785 & 0.575 & 0.410 \\
TPOT          & 63.0  & 66.5  & 82.6  & 84.2  \\
Ratio         & 1.00$\times$ & 1.06$\times$
              & 1.31$\times$ & 1.34$\times$ \\
\bottomrule
\end{tabular*}

\vspace{0.15em}

\begin{tabular*}{\columnwidth}{
@{\extracolsep{\fill}} c c c c c c c c c @{}
}
\toprule
\multicolumn{9}{c}{(b) Sensitivity under natural settings} \\
\midrule
\multicolumn{3}{c}{\textbf{Task Domain}} &
\multicolumn{3}{c}{\textbf{Context Length}} &
\multicolumn{3}{c}{\textbf{Decoding Strategy}} \\
\cmidrule(lr){1-3}
\cmidrule(lr){4-6}
\cmidrule(lr){7-9}
\textbf{Set.} & \textbf{Recall} & \textbf{TPOT} &
\textbf{Set.} & \textbf{Recall} & \textbf{TPOT} &
\textbf{Set.} & \textbf{Recall} & \textbf{TPOT} \\
\midrule
WT & 0.930 & 57.5 &
8 & 0.951 & 70.6 &
Greedy & 0.945 & 68.4 \\

C4 & 0.961 & 78.1 &
16 & 0.949 & 69.0 &
$T=0.7$ & 0.950 & 65.3 \\

GK & 0.954 & 72.7 &
32 & 0.946 & 68.7 &
$T=1.0$ & 0.950 & 74.6 \\
\bottomrule
\end{tabular*}

\endgroup
\end{table}

{{\subsection{Routing-Prediction Sensitivity and Robustness}
\label{ssec:robustness}

Table~\ref{tab:robustness} evaluates sensitivity to routing
mispredictions and robustness across inference conditions.
We replace 0\%, 25\%, 50\%, and 75\% of oracle expert identities
used for prefetching with experts sampled from the same layer,
leaving target routing and speculative verification unchanged.
Since replacements may still select required experts, we report
the resulting accuracy after injection. The zero-error case is
an idealized prefetch baseline that can achieve lower TPOT than
prefetching with Specter's own predictions.
As accuracy decreases from 1.000 to 0.410, TPOT increases from
63.0 to 84.2\,ms/token, a 1.34$\times$ slowdown.

Recall is the fraction of distinct target-selected experts also
selected by the draft per speculative iteration, averaged across
MoE layers and iterations. It remains within 0.930--0.961 across
prompt domains, context lengths, and decoding strategies.
Target verification preserves the output distribution, limiting
the impact of routing mispredictions to prefetching and execution
performance.
}}

\section{Related Work}

\subsection{Efficient MoE Inference Systems}
MoE inference research~\cite{liu2026survey,liu2025optimizing,kamahori2024fiddler, wang2026asymserve}
addresses the memory and communication costs of large parameter
footprints and data-dependent expert activation.
General-purpose frameworks~\cite{wolf-etal-2020-transformers,aminabadi2022deepspeed,kwon2023efficient}
support offloading, while specialized systems optimize expert placement
and offloading~\cite{eliseev2023fast,sarkar2023edge,xue2024moe,tairin2025emoe,zhou2025floe,zhong2025hybrimoe},
caching, and prefetching~\cite{zhong2024adapmoe,tang2026moe,song2024promoe,hwang2024pre}
using routing locality or predicted demand.
Their overlap remains limited when demand is revealed only during
target execution.

\subsection{Speculative Decoding for LLMs}
Work on speculative decoding~\cite{leviathan2023fast,chen2023accelerating}
improves draft training~\cite{zafrir2024fastdraft,wen2024speculative,zhou2023distillspec,hu2025griffin,weng2025coral},
draft trees~\cite{du2024glide,li2024eagle,wang2025opt},
verification~\cite{sun2023spectr,wu2025tetris},
runtime efficiency~\cite{chen2025spin,zhang2025swiftspec},
and dynamic speculation depth~\cite{mamou2024dynamic,brown2024dynamic},
which balances draft overhead against accepted-token gains.

\begingroup
\emergencystretch=2em
For MoE, EARTH~\cite{liu2026earth} and MoE-SpAc~\cite{li2026moe}
use custom hardware or heterogeneous CPU--GPU scheduling;
SP-MoE~\cite{chen2025sp} requires structurally compatible draft--target
pairs without cross-model weight sharing;
SpecMoEOff~\cite{wang2025accelerating} emphasizes speculative I/O pipelining;
and MoE-Spec~\cite{mcdanel2026moe} and SS-MoE~\cite{10.1145/3774904.3792218}
trade quality for efficiency through restricted or approximate expert execution.
\par
\endgroup

\textsc{Specter} couples speculative decoding with expert offloading,
using a quantized draft's multi-token routing lookahead for prefetching
while standard target verification preserves the output distribution.

\subsection{Quantization for LLM Inference}
\begingroup
\emergencystretch=2em
Post-training and mixed-precision quantization reduce LLM memory
and computation costs.
\mbox{SmoothQuant}~\cite{xiao2023smoothquant}, AWQ~\cite{lin2024awq}, GPTQ~\cite{frantar2022gptq}
and other low-bit methods~\cite{park2025decdec,zhao2024atom,chen2025bitmod,liu2025vq}
preserve model quality at reduced precision, while
Marlin~\cite{10.1145/3710848.3710871} accelerates inference through
efficient low-bit GEMMs.
Aggressive quantization can nevertheless degrade generation quality.
\par
\endgroup
\section{Conclusion}

In this paper, we presented \textsc{Specter}, a speculative expert-offloading system for memory-constrained, single-GPU MoE inference. Using a quantized draft model, \textsc{Specter} predicts expert routing across multiple future tokens and converts this lookahead into inter-token expert prefetching. It coordinates staged prefetching, verification-time expert residency, expert-wise execution, and fused draft computation to overlap PCIe transfers with computation. Evaluations demonstrate improved decoding latency and throughput across the evaluated models, workloads, memory budgets, and batch sizes. The system design of \textsc{Specter} allows speculative lookahead to improve execution performance without changing the target model's output distribution.

\begin{acks}
We sincerely thank the anonymous reviewers and our shepherd
for their valuable comments and feedback.
This work is supported by the National Natural Science
Foundation of China under Grant Nos.~62502302 and U23A6007.
\end{acks}

\bibliographystyle{ACM-Reference-Format}
\bibliography{refs}

\appendix
\providecommand{\SpecterArtifactDOI}{\url{https://doi.org/10.5281/zenodo.22937478}}
\providecommand{\SpecterArtifactPublic}{Yes}
\begingroup
\setlength{\emergencystretch}{1em}

\section{Artifact Appendix}
\label{sec:specter-artifact}

\subsection{Abstract}
Specter uses speculative expert offloading for MoE inference on a single
GPU with limited memory. This artifact provides installation instructions
and scripts for decode latency, baseline comparisons, speculative depth,
input length, and system measurements. The supplied DeepSeek-V2-Lite
checkpoints support high- and low-memory experiments with Specter,
SpecMoEOff, and Mixtral-Offloading. Qwen1.5-MoE and Phi-3.5-MoE are also
supported with separately prepared checkpoints. The
\href{https://github.com/wwwwwf/specter_asplos}{GitHub repository}
provides the source, checkpoint downloads, and complete instructions.

\subsection{Artifact check-list (meta-information)}
\begin{itemize}
  \setlength{\itemsep}{0pt}
  \setlength{\parskip}{0pt}
  \item \textbf{Algorithm:} Speculative expert offloading for MoE inference.
  \item \textbf{Run-time environment:} Linux x86-64, Python 3.11.13,
    PyTorch 2.7.0 with CUDA runtime 12.6, and CUDA Toolkit 12.8.
  \item \textbf{Hardware:} One NVIDIA A100 GPU with 40\,GB memory;
    the reference machine has 256\,GB host memory.
  \item \textbf{Disk space required:} Approximately 200\,GB for downloads,
    extracted checkpoints, the environment, and caches.
  \item \textbf{Time needed to prepare workflow:} About one hour,
    excluding downloads.
  \item \textbf{Time needed to complete experiments:} Approximately one
    hour for main, depth, and prefix; baseline comparisons and other
    measurements require additional time.
  \item \textbf{Publicly available?:} \SpecterArtifactPublic
  \item \textbf{Archived DOI?:}\par
    \SpecterArtifactDOI
\end{itemize}

\subsection{Description}
\subsubsection{How to access.}
The archived source corresponds to Git commit \texttt{c0325dc}.
Download \path{specter-ae-v1.tar.gz}, \path{specter-ae-wheels.tar.gz},
\path{specter-ae-datasets.tar.gz}, \path{SHA256SUMS}, and
\path{specter-ae-v1.manifest.json} from the Zenodo record.
Choose a workspace with sufficient free space and place these five files
in its \path{downloads/} directory:
\begin{Verbatim}[fontsize=\scriptsize,frame=single,framesep=3pt,samepage=true]
export SPECTER_WORK=/path/to/specter-work
mkdir -p "$SPECTER_WORK/downloads" \
  "$SPECTER_WORK/logs"
\end{Verbatim}
Verify the files, extract the source, and download the prepared
checkpoints from the
\href{https://github.com/wwwwwf/specter_asplos/releases/tag/ae-models-20260927}{DeepSeek release}:
\begin{Verbatim}[fontsize=\scriptsize,frame=single,framesep=3pt,samepage=true]
cd "$SPECTER_WORK/downloads"
sha256sum --ignore-missing -c SHA256SUMS
cd ..
tar -xzf downloads/specter-ae-v1.tar.gz
cd specter
python3 -B initialization/download_models.py \
  --manifest initialization/model-assets.json \
  --output "$SPECTER_WORK/downloads"
cd ../downloads
sha256sum -c SHA256SUMS
cd ..
for f in wheels datasets dsv2-target dsv2-draft
  do tar -xzf "downloads/specter-ae-$f.tar.gz"
  done
unzip -n -P deserted-untie-orchid \
  specter-assets/datasets/gpqa/gpqa_main.csv.zip \
  -d specter-assets/datasets/gpqa/
\end{Verbatim}
Require four successful checks initially and six after downloading the
models. Assets are extracted under \path{specter-assets/}; the README
provides complete download commands.

\subsubsection{Hardware dependencies.}
Experiments use one A100 40\,GB GPU at batch size one. The reference
platform has an Intel Xeon Silver 4310 CPU and 256\,GB host memory
for offloaded experts. NUMA binding is optional.

\subsubsection{Software dependencies.}
Requires a CUDA driver, CUDA Toolkit 12.8, a C++ compiler,
Python 3.8 or newer for downloading, \texttt{curl}, \texttt{tar}, and
\texttt{unzip}. The supplied wheels are GPTQModel 4.0.0.dev0,
Transformers 4.53.3, FlashAttention 2.8.1, and vLLM 0.9.2.
Use these validated builds; matching upstream version numbers are
insufficient. The installer verifies wheel hashes and installs the
remaining pinned dependencies from an online package index.

\subsubsection{Expected runtime.}
The reference main run takes about 20 minutes; each depth or prefix sweep
takes about 20--30 minutes, including model loading. Baseline comparisons
run three methods per memory setting. Runtime depends on the host and
storage system.

\subsection{Installation}
Install \texttt{uv} without \texttt{sudo} and place caches and temporary
build files on the work filesystem:
\begin{Verbatim}[fontsize=\scriptsize,frame=single,framesep=3pt,samepage=true]
cd "$SPECTER_WORK/specter"
set -o pipefail
curl -LsSf \
  https://astral.sh/uv/0.12.19/install.sh | \
  env UV_INSTALL_DIR="$HOME/.local/bin" sh
export PATH="$HOME/.local/bin:$PATH"
export UV_CACHE_DIR="$SPECTER_WORK/cache/uv"
export UV_PYTHON_INSTALL_DIR=\
"$SPECTER_WORK/cache/python"
export TMPDIR="$SPECTER_WORK/cache/tmp"
mkdir -p "$UV_CACHE_DIR" \
  "$UV_PYTHON_INSTALL_DIR" "$TMPDIR"
cp config.example.toml ../specter.local.toml
export SPECTER_CONFIG=\
"$SPECTER_WORK/specter.local.toml"
uv run --python 3.11.13 --no-project \
  environment/install.py \
  --config "$SPECTER_CONFIG"
source "$SPECTER_WORK/env/bin/activate"
\end{Verbatim}
The copied configuration matches the supplied assets. Relative paths
resolve from its directory; edit them when reusing other assets.
The environment, caches, and results remain outside the source tree.
The installer pins \texttt{flashinfer-python} and applies
\path{environment/build-constraints.txt}, including
\texttt{torch==2.7.0}, to its isolated build.

Select an A100, set \texttt{CUDA\_HOME} to the local CUDA 12.8 toolkit,
validate resources, build kernels, and run the tests:
\begin{Verbatim}[fontsize=\scriptsize,frame=single,framesep=3pt,samepage=true]
export CUDA_DEVICE_ORDER=PCI_BUS_ID
export CUDA_VISIBLE_DEVICES=0
export CUDA_HOME=/usr/local/cuda-12.8
export TORCH_CUDA_ARCH_LIST=8.0 MAX_JOBS=2
export PYTHONNOUSERSITE=1 PYTHONUNBUFFERED=1
export PYTHONDONTWRITEBYTECODE=1
python -B -m configuration.check \
  --config "$SPECTER_CONFIG"
python -B -m \
  streamlined_execution_engine.kernels.build
python -B -m pytest -p no:cacheprovider \
  tests baselines -q
\end{Verbatim}
Continue after all checks succeed. The resource check validates
checkpoint files and reads five nonempty inputs from each dataset.

\balance
\subsection{Alternate Setup: Existing Environment}
Update the source files to this release and reuse the installed
environment, configuration, datasets, and DeepSeek checkpoints.
Activate the environment, export \texttt{SPECTER\_CONFIG}, and select
the GPU and toolkit as above. The original main, depth, and prefix
commands retain their workloads and parameters.

\subsection{Accessing DeepSeek-V2-Lite}
The public release supplies the target checkpoint, loaded in FP16,
and a draft with routed experts quantized to GPTQ INT4 (group size 128).
No local quantization is needed. The download helper verifies and
reconstructs the split archives and resumes interrupted downloads;
GitHub login is not required. The five supplied datasets are GSM8K,
WikiText-2, HumanEval, GPQA, and C4, with configuration keys
\texttt{GK}, \texttt{WT}, \texttt{HE}, \texttt{GP}, and \texttt{C4}.

\subsection{Experiment workflow}
\label{sec:specter-ae-workflow}
Keep \texttt{SPECTER\_CONFIG} exported. Run the default high-memory
main experiment and supporting sweeps:
\begin{Verbatim}[fontsize=\scriptsize,frame=single,framesep=3pt,samepage=true]
python -B -m benchmarks.run_ae
python -B -m benchmarks.run_ae --experiment depth
python -B -m benchmarks.run_ae --experiment prefix
\end{Verbatim}
Use \texttt{--experiment all} to run these three sequentially.
The \texttt{--memory} option accepts \texttt{high}, \texttt{low}, or
\texttt{both} for \texttt{main}, \texttt{all}, and \texttt{speedup}.
Run the low-memory main experiment or compare Specter with both
baselines in both memory settings:
\begin{Verbatim}[fontsize=\scriptsize,frame=single,framesep=3pt,samepage=true]
python -B -m benchmarks.run_ae \
  --experiment main --memory low --output ae_low
mkdir -p "$SPECTER_WORK/logs"
set -o pipefail
python -B -m benchmarks.run_ae \
  --experiment speedup --memory both \
  --output ae_compare \
  2>&1 | tee "$SPECTER_WORK/logs/ae_compare.log"
\end{Verbatim}
The baseline implementations are in \path{baselines/}. High/low
resident counts per layer are 16/4 for Specter, 12/2 for SpecMoEOff,
and 32/20 for Mixtral-Offloading. Specter uses $K=16$; SpecMoEOff
uses $K=3$. Routing and sensitivity measurements use:
\begin{Verbatim}[fontsize=\scriptsize,frame=single,framesep=3pt,samepage=true]
python -B -m benchmarks.run_ae --experiment routing
python -B -m benchmarks.run_ae \
  --experiment sensitivity
\end{Verbatim}
Other experiment names are \texttt{working-set}, \texttt{cache},
\texttt{ablation}, \texttt{latency}, \texttt{memory}, \texttt{fidelity},
\texttt{oracle}, and \texttt{kernel}. Add \texttt{--dry-run} to inspect
commands before execution.

\subsection{Evaluation and expected results}
Results are saved under \path{paths.output}, which defaults to
\texttt{\$SPECTER\_WORK/results}. The main run writes
\path{records.jsonl}, \path{summary.json}, and \path{summary.csv}
under \path{ae/main/}; the low-memory command above writes
\path{ae_low/main_low/}. The reference high-memory main run achieves
approximately 70\,ms/token. Depth profiling writes to \path{ae/depth/}
and selects the lowest-TPOT $K$. Prefix results are under
\path{ae/prefix_*/}.

The comparison command writes \path{comparison.json} and
\path{comparison.csv} under \path{ae_compare/speedup_high/} and
\path{ae_compare/speedup_low/}, with per-method records below each.
Tables report per-dataset and aggregate TPOT, speedup, output agreement,
and GPU memory. Speedup is baseline mean decode TPOT divided by
Specter mean decode TPOT. The external log records progress and results.

Other measurements write \texttt{ae/<experiment>/}. Generate plots
and source tables using the corresponding result directory:
\begin{Verbatim}[fontsize=\scriptsize,frame=single,framesep=3pt,samepage=true]
python -B -m benchmarks.plot_experiments \
  --input "$SPECTER_WORK/results/ae/routing"
\end{Verbatim}
Plots are saved in its \path{figures/} subdirectory.

\subsection{Experiment customization}
Main and baseline comparisons use greedy decoding, five inputs from
each of five datasets, three repeats, a 16-token prefix cap, and 128
output tokens. Each method produces 75 records per memory setting.
Depth and prefix sweeps use three GSM8K inputs, three repeats, and
128 output tokens. Depth tests $K\in\{2,4,6,8,12,16,20,24,32\}$;
prefix tests caps of 8, 16, and 32 tokens. See the README for custom
workloads. Use a fresh \texttt{--output} name for another run;
relative names resolve under \path{paths.output}.
\par
\endgroup

\end{document}